\documentclass[runningheads]{llncs}
\usepackage{graphicx}
\usepackage{multirow}
\usepackage{lscape}
\usepackage{siunitx}
\usepackage{amsmath}
\usepackage{subcaption}
\usepackage{wrapfig}

\def\mat{\textsc{Mat}}
\def\matstart{\textsc{Mat-Start}}
\def\matfinal{\textsc{Mat-Final}}
\def\matsolvent{\textsc{Mat-Solvent}}
\def\ope{\textsc{Ope}}
\def\property{\textsc{Prop}}
\def\propertytime{\textsc{Prop-Time}}
\def\propertytemp{\textsc{Prop-Temp}}
\def\propertyequipment{\textsc{Prop-Equip}}
\def\propertymanufacturer{\textsc{Prop-Maker}}
\def\propertymethod{\textsc{Prop-Method}}
\def\characteristic{\textsc{Chara}}
\def\characteristicname{\textsc{Chara-Name}}
\def\characteristicconductivity{\textsc{Chara-Cond}}
\def\characteristicactivation{\textsc{Chara-Act}}

\newcommand{\tref}[1]{Tab.~\ref{#1}}
\newcommand{\Tref}[1]{Table~\ref{#1}}

\newcommand{\Fref}[1]{Figure~\ref{#1}}

\newcommand{\Sref}[1]{Section~\ref{#1}}

\begin{document}
\title{Analyzing Research Trends\\in Inorganic Materials Literature Using NLP}
\titlerunning{Analyzing Research Trends in Literature}

\author{Fusataka Kuniyoshi\inst{1,3} \and
Jun Ozawa\inst{1,3} \and
Makoto Miwa\inst{2,3}}

\authorrunning{F. Kuniyoshi et al.}

\institute{
Panasonic Corporation, 1006, Oaza Kadoma, Kadoma-shi, Osaka 571-8501, Japan
\and
Toyota Technological Institute, 2--12--1, Hisakata, Tempaku-ku, Nagoya, 468--8511, Japan
\and
National Institute of Advanced Industrial Science and Technology (AIST), 2--3--26, Aomi, Koto-ku, Tokyo, 135--0064, Japan
\\
\email{\{kuniyoshi.fusataka, ozawa.jun, makoto.miwa\}@aist.go.jp}
}

\maketitle              
\begin{abstract}
In the field of inorganic materials science, there is a growing demand to extract knowledge such as physical properties and synthesis processes of materials by machine-reading a large number of papers. This is because materials researchers refer to many papers in order to come up with promising terms of experiments for material synthesis. However, there are only a few systems that can extract material names and their properties. This study proposes a large-scale natural language processing (NLP) pipeline for extracting material names and properties from materials science literature to enable the search and retrieval of results in materials science. Therefore, we propose a label definition for extracting material names and properties and accordingly build a corpus containing 836 annotated paragraphs extracted from 301 papers for training a named entity recognition (NER) model. Experimental results demonstrate the utility of this NER model; it achieves successful extraction with a micro-F1 score of 78.1\%. To demonstrate the efficacy of our approach, we present a thorough evaluation on a real-world automatically annotated corpus by applying our trained NER model to 12,895 materials science papers. We analyze the trend in materials science by visualizing the outputs of the NLP pipeline. For example, the country-by-year analysis indicates that in recent years, the number of papers on ``MoS$_2$,'' a material used in perovskite solar cells, has been increasing rapidly in China but decreasing in the United States. Further, according to the conditions-by-year analysis, the processing temperature of the catalyst material ``PEDOT:PSS'' is shifting below 200 $^{\circ}$C, and the number of reports with a processing time exceeding 5 h is increasing slightly.

\end{abstract}
\section{Introduction}
Materials science literature includes considerable information such as material names and their properties described in natural language. Therefore, the automatic extraction of the details necessary to reproduce and validate materials synthesis processes in a materials science laboratory remains difficult and requires extensive human intervention. The automatic compilation of such literature into a structured form could enable realizing a data-driven materials discovery system that does not require human intervention; such a system could become a key enabler in the design and discovery of novel materials. In this regard, named entity recognition (NER) is helpful, which seeks to locate spans and classify named entities in unstructured text into predefined categories such as material name.

NER has already found many applications in materials science. For example, material names have been linked to their properties, such as characteristic values or their structures, through a combination of database lookup and the parsing of systematic nomenclature to create reader-friendly semantically enhanced literature~\cite{Tshitoyan2019UnsupervisedWE,Kononova2019TextminedDO}. Further, NER has been linked to material information retrieval techniques to search for materials similar to a query material from corpora~\cite{Kim2017MaterialsSI,Kim2017MachinelearnedAC,Young2017DataMF,Huang2020ADO}, or to predict the characteristic values of a query material~\cite{Jensen2019AML,Court2020MagneticAS}.

A technique that can extract natural language characteristic values and link a material name to a machine-readable representation will find importance in many practical applications. We believe that current research in this area is hampered by the lack of available annotated corpora.

\begin{figure}[t]
 \centering
 \includegraphics[width=\linewidth]{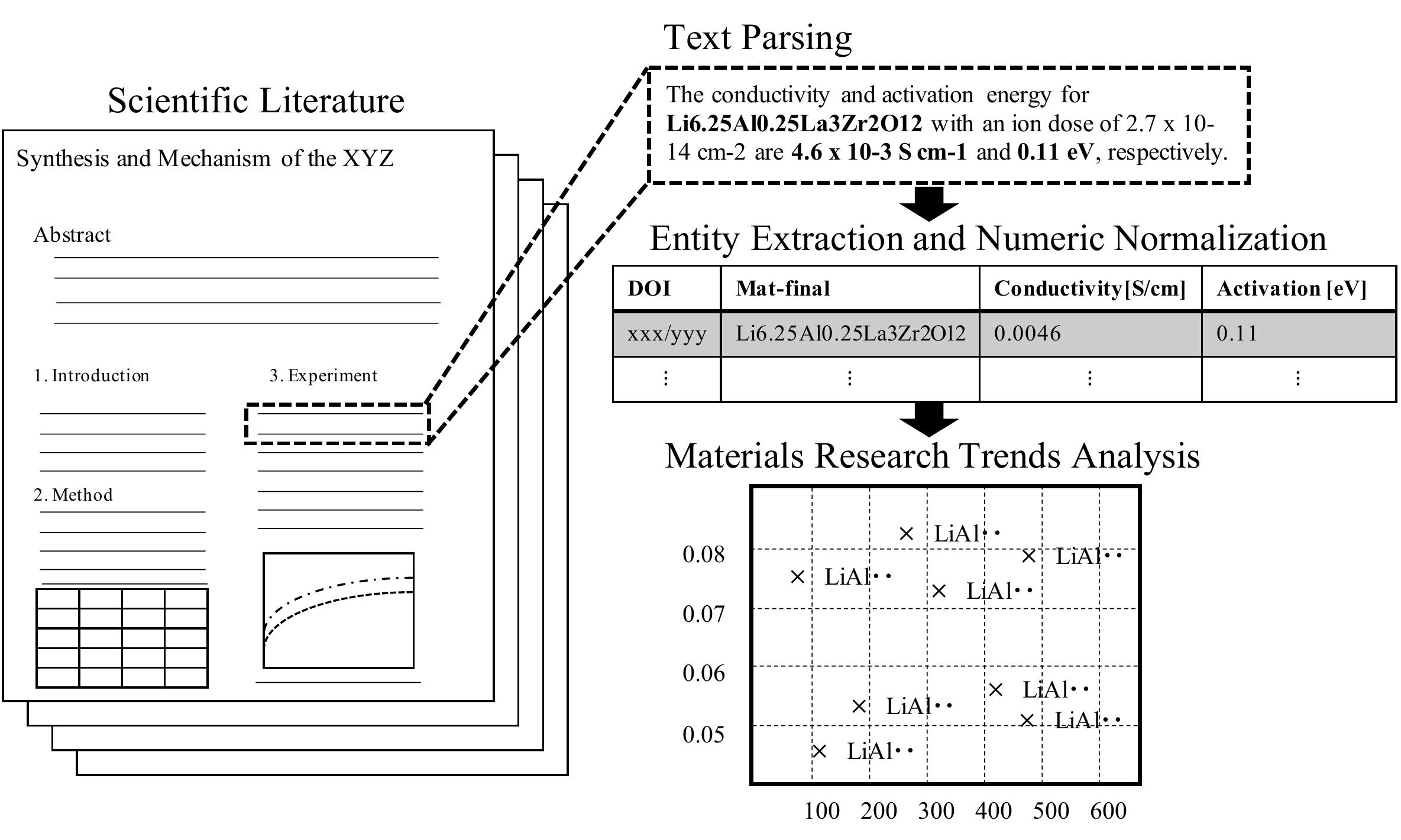}
 \caption{Overview of pipeline for extraction.}
 \label{fig:overview}
\end{figure}

In this study, we propose a natural language processing (NLP)-based approach to analyze the trend in materials for development in materials science (see \Fref{fig:overview}).
Toward this end, we propose a pipeline that integrates an NER model and a numeric normalization module. To evaluate this pipeline, we annotate 836 paragraphs extracted from 301 papers to extract material terminology and conduct initial analyses to extract material data from 12,895 unlabeled full-text literature. Through this evaluation, we demonstrate the reliability of our NLP framework by presenting the detailed NER model training process and by showing the detailed evaluation of the trained NER model. Our NER model can extract material names and several important properties such as temperature, time, conductivity, and activation energy. To demonstrate the utility of our annotated corpus and analyze the research trends, we explore the extracted outputs of our NLP framework from 12,895 unlabeled materials science literature. This study makes the following contributions:

\begin{itemize}
    \item We propose a manually annotated corpus and an NLP framework for extracting material names with properties using an NER tagger and apply a numeric normalization module to NER outputs.
    \item We evaluate the reliability of the NLP system using by showing the detailed process of training the NER model with sufficient evaluation.
    \item We demonstrate the analysis for observing the research trends using our NLP outputs for material knowledge discovery from scientific literature. 
\end{itemize}

\section{Related Work}
Many NLP systems and language resources are available for extracting different types of information from scientific literature: identifying drug names~\cite{Krallinger2015}, discovering drugs~\cite{Hansson2020SemanticTM}, examining the side-effects of drugs~\cite{Jeong2020ExaminingDA}, extracting biomedical terminology~\cite{Bada2011ConceptAI,RebholzSchuhmann2012TextminingSF,Weber2021HunFlairAE} or events~\cite{Kim2003GENIAC,Miwa2012BoostingAE,Bjrne2018BiomedicalEE}, and extracting wet-lab protocol~\cite{kulkarni-etal-2018-annotated} are some of these examples.

In inorganic materials science, text mining is mainly used to search for a domain-specific material name or for classifying materials by their type: inorganic materials in general~\cite{Kim2017MaterialsSI,Young2017DataMF,doi:10.1021/acs.jcim.9b00995,Kononova2019TextminedDO,Huo2019SemisupervisedMC,mysore-etal-2019-materials}, oxides~\cite{Kim2017MachinelearnedAC}, superconductors~\cite{Yamaguchi2020SCCoMIcsAS,Court2020MagneticAS}, zeolites~\cite{Jensen2019AML}, and battery materials~\cite{Huang2020ADO,Kuniyoshi2020,Mahbub2020TextMF}.

However, in inorganic materials science, few practical systems have been proposed to extract material names from a large number of papers by associating them with their property values. In this study, we propose an NLP system for extracting material names and properties.

\section{Corpus Preparation}

\subsection{Definition of Types}
Our proposed annotation scheme is based on Kuniyoshi's annotation scheme for materials synthesis processes~\cite{Kuniyoshi2020}. We used 12 labels that were defined to annotate spans of text; these represent the materials, operations, and properties. In the list, we segmented the roles of materials (\mat{}), operations (\ope{}), properties (\property{}), and characteristics (\characteristic{}).

\textbf{\matfinal{}} represents the final material (or product) of the material synthesis process; for example, ``A solid solution of the lithium superionic conductor \underline{Li$_{10+}\delta$Ge$_{1+}{\delta}$P$_{2-}{\delta}$S$_{12}$} ($0 \leq \delta \leq 0.35$) was synthesized ...''

\textbf{\matsolvent{}} is a liquid that is used to dissolve substances and create solutions; for example, ``Ga$_{2}$O$_{3}$ (99.999\%) were ground and homogenized in \underline{ethanol}.''

\textbf{\matstart{}} is a raw material used to synthesize the final material; for example, ``Precursor powders (10 g) containing a stoichiometric mixture of \underline{La$_{2}$O$_{3}$} (99.998\%).''

\textbf{\ope{}} represents an individual action performed by the experimenters. It is often represented by verbs; for example, ``Carbon black was \underline{dried} at 80$^{\circ}$C.''

\textbf{\propertyequipment} represents equipment for analyzing a material; for example, ``... spectrum analysis of the films was carried out on a \underline{UV-Vis spectrophotometer}.''

\textbf{\propertymanufacturer} represents a manufacturer of equipment or material powder; for example, ``m-Cresol was obtained from \underline{Sigma-Aldrich}.''

\textbf{\propertymethod} represents a method to analyze a material sample; for example, ``The surface morphologies of the relevant membranes were studied by using a high-resolution \underline{field-emission scanning electron microscopy}.''

\textbf{\propertytemp{}} represents a temperature condition associated with an operation; for example, ``... finally dried at \underline{80$^{\circ}$C} in vacuum for 5 h.''

\textbf{\propertytime{}} represents a time condition associated with an operation; for example, ``... finally dried at 80$^{\circ}$C in vacuum for \underline{5 h}.''

\textbf{\characteristicname} represents a characteristic name to classify characteristic values; for example, ``... the glass ceramic has a room-temperature \underline{ionic conductivity} as high as 3 $\times$ 10$^{-5}$ S cm$^{-1}$.''

\textbf{\characteristicactivation} represents a characteristic value of activation energy. For example, the unit of \textbf{activation energy} is eV; then, ``The activation energy as function of the vacancy concentration exhibits a minimum of \underline{0.7 eV} ...''

\textbf{\characteristicconductivity} represents a characteristic value of conductivity. For example, the unit of \textbf{conductivity} is S/cm; then, ``The ionic conductivity of the prepared pellets is \underline{1.03 $\times$ 10$^{-3}$ S/cm}.''

\subsection{Collecting Literature}
Our corpus was constructed from papers published in a journal---Journal of Material Chemistry A (JMCA; Royal Society of Chemistry (RSC))\footnote{\url{https://www.rsc.org/journals-books-databases/about-journals/journal-of-materials-chemistry-a/}}---from 2015 to 2019. JMCA focuses on energy and sustainability, and it publishes papers discussing materials such as solar cells, thermoelectric conversion materials, liquid lithium ion batteries (LIBs), and all-solid-state batteries. The RSC provides papers in XML format, in which contents have a hierarchical structure within different nested tags. For example, the $<$section$>$ tag contains information such as the section title and paragraph. To extract plain text from such XML files, we created an extraction tool that exploits RSC's semantic markup features to extract information such as the title, abstract, and main contents. Then, we stripped out the embedded markups to produce the plain text and to create a linear stream of elements containing all data in the papers. These text data were then transferred into a document object comprising subobjects such as title, heading, and paragraph. Further, we automatically extracted paragraphs with their section and subsection titles by using regular expressions for target titles such as ``Abstract,'' ``Introduction,'' ``Experimental,'' and ``Conclusion.''

\subsection{Annotation}
\begin{figure}[t]
\centering
    \includegraphics[width=\linewidth]{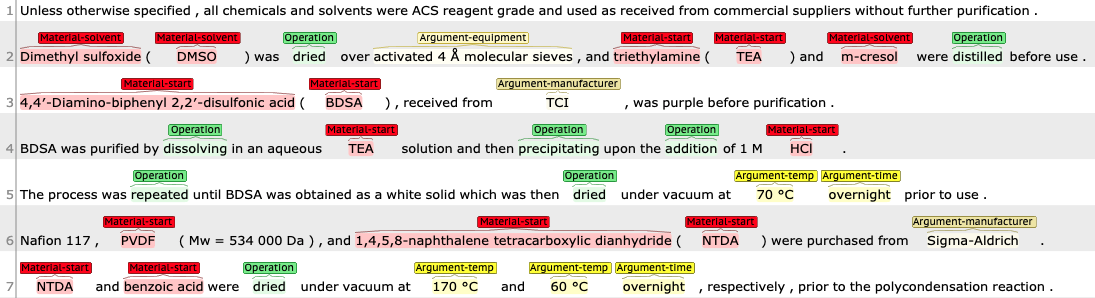}
    \caption{Example of annotation of experimental section. Text labeling interface to annotate material names and their properties. Example of annotation of experimental sections. This texts is referred from the study by Yuan~\cite{C4TA04910A}.}
  \label{fig:brat}
\centering
\end{figure}

One Master's degree staff in the materials science annotated labels on the 836 paragraphs extracted from 301 papers. \Fref{fig:brat} illustrates annotations to the text in the experimental section, by using the brat annotation toolkit~\cite{Stenetorp2012bratAW}. The annotated data are converted into the Inside, Outside, Beginning (IOB) scheme, where a token is labeled as \textit{I-$\ast$} if it is inside a named entity of type $\ast$, \textit{O} if it is outside of named entities, and \textit{B-$\ast$} if it is at the beginning of an $\ast$ entity. Therefore, the model is trained to classify each word in a sequence into 25 different labels consisting of one \textit{O} label and \textit{B-$\ast$} and \textit{I-$\ast$} labels for each of the 12 entity labels. Our corpus is shared at github repository\footnote{\url{https://github.com/BananaTonic/Material\_Synthesis\_Corpus.git}}.

\section{Approach}
\begin{figure}[t]
 \centering
 \includegraphics[width=\linewidth]{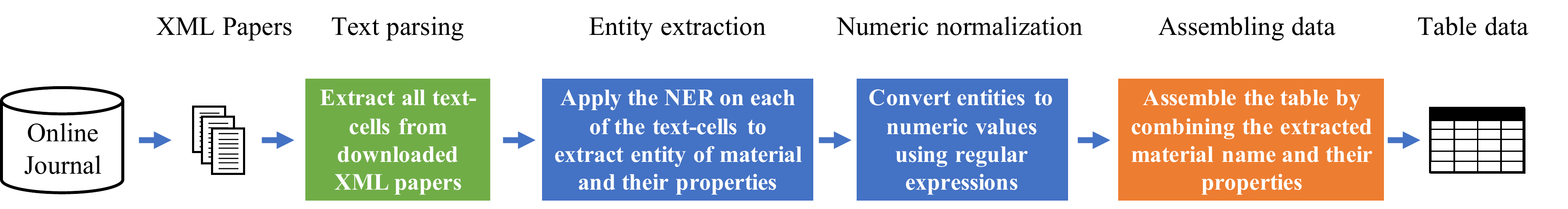}
 \caption{Overview of analysis pipeline.}
 \label{fig:pipeline}
\end{figure}

This section explains our framework for extracting material data, such as names and property values, from a large number of papers. Our framework (see \Fref{fig:pipeline}) consists of an NER-based sequence labeling tool and a module that converts a natural language phrase to numeric values.

\subsection{Sequence Labeling Architecture}
First, we briefly describe bidirectional long short-term memory (BiLSTM), a type of recurrent neural network, and a subsequent conditional random field (CRF). Then, we explain the hybrid labeling architecture that is based on a previous study~\cite{Akbik2018ContextualSE,Huang2015BidirectionalLM}.

\begin{wrapfigure}{r}[10pt]{0.45\textwidth}
 \centering
 \includegraphics[width=\linewidth]{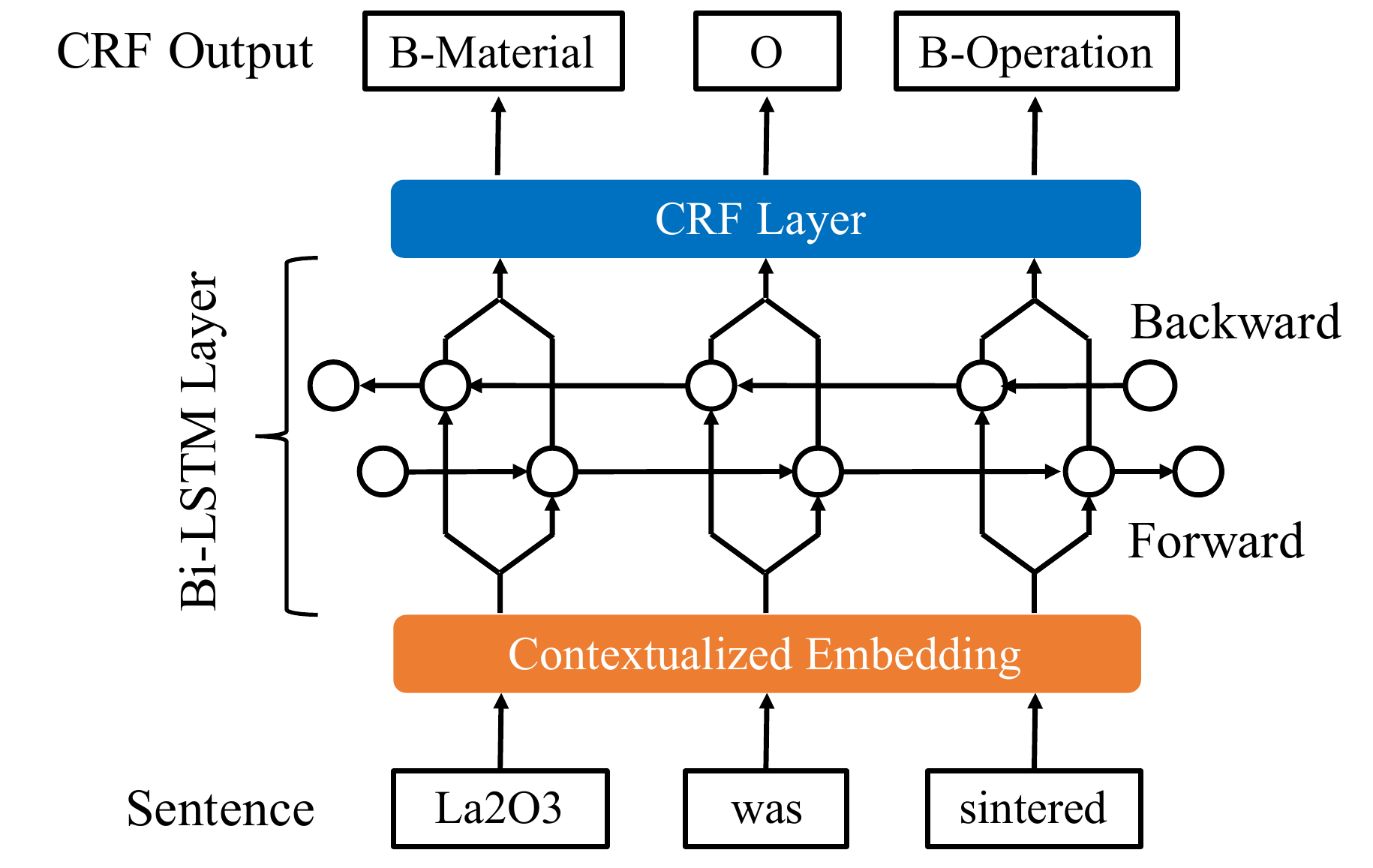}
 \caption{Proposed NER architecture.}
 \label{fig:ner}
\end{wrapfigure}

We extract the output hidden state after and before the word's token in a sentence from the corresponding forward and backward LSTMs to capture semantic-syntactic information from the beginning and ending of the sentence to the token, respectively. Both output hidden states are concatenated to form the final embedding and to capture the semantic-syntactic information of the word itself as well as its surrounding context. \Fref{fig:ner} shows our proposed NER architecture.

Let the individual tokens in a sentence be $\it{t_0, t_1, ..., t_n}$. We define the contextual string embeddings of these tokens as $\it{h_0, h_1, ..., h_n}$, where $h_t$ represents the output hidden state of a token $t$. The final word embeddings are passed to a BiLSTM-CRF sequence labeling module to address downstream sequence labeling tasks. 

Calling the inputs to the BiLSTM gives
\begin{equation*}
    \mathbf{r}_i \simeq [ \mathbf{r}^f_i; \mathbf{r}^b_i ],
\end{equation*}
where $\mathbf{r}^f_i$ and $\mathbf{r}^b_i$ are forward and backward output states of the BiLSTM, respectively. The final sequence probability is then given by a CRF over the possible sequence labels $\mathbf{y}$:
\begin{equation*}
    \hat{P}(\textbf{y}|\textbf{r}) \propto \prod_{i=1}^n \phi(\textbf{y}, \textbf{r}),
\end{equation*}
where $\phi(\cdot)$ is a variation Markov Random Field of all clique potentials. Finally, the prediction of the label is given by
\begin{equation*}
    P(\textbf{y}_i = j|\textbf{r}_i) = softmax(\textbf{r}_i)[j]
\end{equation*}

\subsection{Numeric Normalization}
\begin{figure}[t]
 \centering
 \includegraphics[width=\linewidth]{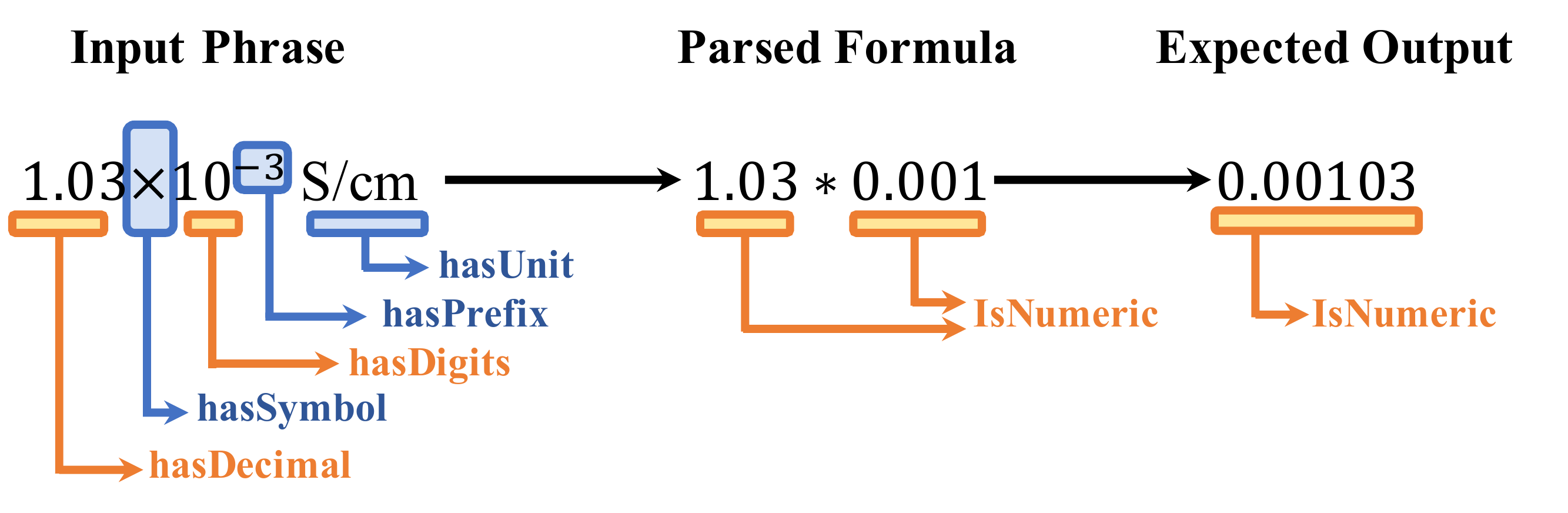}
 \caption{Example of numeric normalization using pattern matching.}
 \label{fig:reg}
\end{figure}

We normalize the numeric values in a post-processing step.
Although the phrases of the characteristic values extracted by the aforementioned named entity extractor represent numerical values, they are annotated in various ways. Therefore, they were normalized into a unified format that allows for comparisons and statistical processing. For example, when ``1.03$\times$10$^{-3}$ S/cm'' was extracted, it was normalized to a value of 0.00103 [S/cm]. In the text of the RSC paper, ``10$^{-3}$'' was described as ``10$<$sup$>$-3$<$/sup$>$.’’ However, in this study, the XML tag was removed beforehand for simplicity, and the text was converted to the plain text ``10 -3.'' In the case of this string, the string ``0.1'' was extracted as a value, ``$\times$'' was extracted as a multiplication sign, and ``10 -3'' was extracted as the 3rd negative exponent of 10. Then, the extracted numbers were multiplied and normalized into the value ``0.00103.'' \Fref{fig:reg} shows an example of our numeric normalization, and \Tref{tab:trans_characteristic} shows the string patterns that can be normalized by this system and their normalization results. The practical workflow of our numeric normalization is as follows: first, when we find measurement patterns in texts, we separate specific expressions into the numeric and unit parts. For example, when ``14 $^\circ$C -- \underline{room temperature}'' was found, we replaced it with ``14 $^\circ$C -- \underline{22 $^\circ$C}.'' Next, we split extracted phrases into specific units such as ``S/cm'' and ``$^\circ$C.'' For example, when ``irradiation times of 3 \underline{s} to 8 \underline{min}'' was extracted, we split this as ``[``irradiation times of 3'', ``to 8'']'' and ``[``s'', ``min'']''. Finally, we extracted each value from split phrases when a unit had numeric patterns before it. For example, when we extracted ``[``0.53-0.58'']'' and ``[``eV'']'' through the previous operation, we extracted the values as the numeric values 0.53 and 0.58 and the unit ``eV.'' In addition, there are variations in the expressions of characteristic values and units in each paper, such as the use of the $\pm$ symbol to indicate conductivity. Therefore, we used regular expressions to write down patterns of values, multiplication symbols, and powers and normalized phrases matching the written patterns into values.

\begin{table}[t]
    \begin{center}
    \begin{tabular}{lcc}\hline
        Type & String pattern & Normalization results with unit \\ \hline
        Temp. & room temperature or RT & 22 $^\circ$C \\
        Temp. & 500 K & 227 $^\circ$C \\
        Time & overnight & 8 hours \\
        Time & half an hour or half a day & 0.5 hours or 12 hours \\
        Time & two hours or 2 h & 2 hours \\
        Cond. & 1.66 $\times{}$ 10$^{-4}$ S/cm & 0.000166 S/cm \\
        Cond. & 4.2 mS/cm & 0.0042 S/cm \\
        Cond. & 4.28 $\pm$ 0.41 $\times$ 10$^{-2}$ & 3.87 S/cm, 4.69 S/cm \\
        AE & 0.93-1.04 & 0.93 eV, 1.04 eV \\
        AE & 2.00(5) eV & 1.95 eV, 2.05 eV \\
        AE & 0.44 $<$ Ea(eV) $<$ 0.46 & 0.44 eV, 0.46 eV \\ \hline \\
    \end{tabular}
    \caption{Example of normalization results for temperature (Temp.), time, conductivity (Cond.), and activation energy (AE). The type indicates the type of phrase, the string pattern indicates the text extracted by the NER extractor, and the normalization result indicates the value normalized to numerical data using regular expressions.}
    \label{tab:trans_characteristic}
    \end{center}
\end{table}

\section{Results}

\subsection{Inter-Annotator Agreement}
\label{sec:iaa}
\begin{figure}
	\begin{minipage}[c]{0.45\textwidth}
	    \begin{center}
		\begin{tabular}{lc} \hline
		    Label & Recall \\ \hline
		    \matstart{} & 0.720  \\
		    \matsolvent{} & 0.529 \\
		    \matfinal{} & 0.697 \\
		    \ope{} & 0.709 \\
		    \propertyequipment{} & 0.537 \\
		    \propertymanufacturer{} & 0.870 \\
		    \propertymethod{} & 0.569 \\
		    \propertytemp{} & 0.722 \\
		    \propertytime{} & 0.792 \\
		    \characteristicname{} & 0.867 \\
		    \characteristicactivation{} & 0.846 \\
		    \characteristicconductivity{} & 0.975 \\
		    ALL & 0.736 \\
		    \hline \\
		\end{tabular}
		\subcaption{Recall of each label. ALL is the overall macro-recall score.}
		\label{tab:iaa_recall}
		\end{center}
	\end{minipage}%
	\begin{minipage}[h]{0.45\textwidth}
		\begin{center}
			\includegraphics[clip, width=1.1\textwidth]{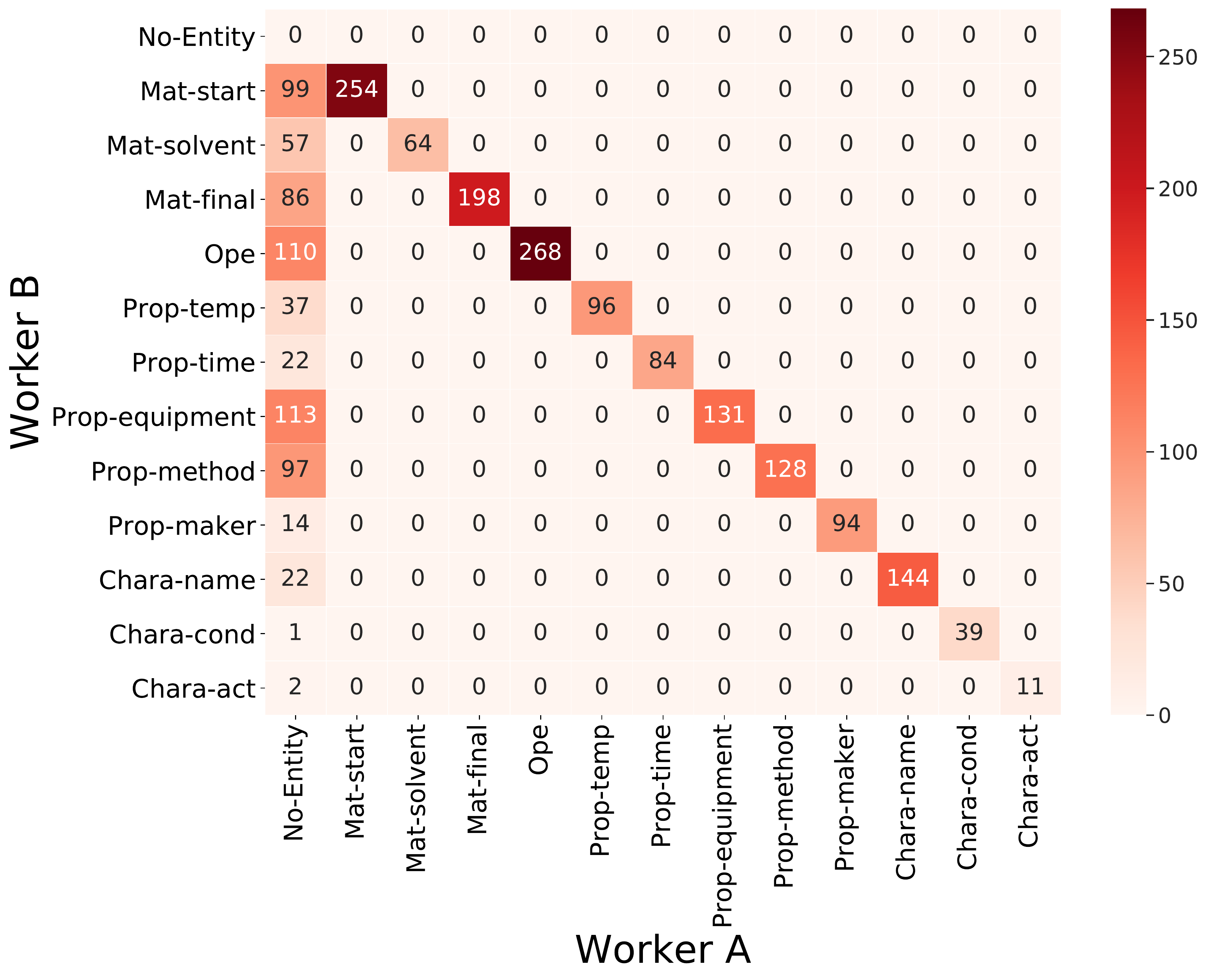}
		\subcaption{Confusion matrix of IAA.}
		\label{fig:iaa_confusion}
		\end{center}
	\end{minipage}
	\caption{IAA results.}
	\label{fig:iaa}
\end{figure}

The inter-annotator agreement (IAA) was evaluated to assess the reliability of the corpus. The IAA is calculated based on the matching of the spans of labels between two annotators who have master's degrees in materials science. The agreement score was calculated by considering the labels identified by one annotator as the gold label and those identified by the other annotator as the prediction. To evaluate the extraction performance, we performed a binary evaluation that classified all entities into either positive or negative. The precision was defined as the fraction of entities predicted as positive that are in fact positive, and recall is defined as the fraction of positive entities that are correctly predicted as positive. More precisely, for true-positive (TP), false-positive (FP), and false-negative (FN) entities, based on the entities extracted by the model, we define precision = TP/(TP + FP), recall = TP/(TP + FN), and F1-score = 2 $\times$ precision $\times$ recall/(precision + recall). These validations were used to evaluate the machine extraction performance when worker A's labeling was considered the correct answer. To verify that the definitions of the types extracted are consistent among the annotators, we used the recall as an evaluation metric. \Tref{tab:iaa_recall} and \Tref{fig:iaa_confusion} show the calculated IAA scores and confusion matrix of each label, respectively, by using 60 paragraphs from 10 papers in our corpus. The result showed that the overall recall of IAA was 0.736, indicating good agreement between the two annotators. The confusion matrix showed that there were no type errors; however, there were many discrepancies owing to misses.

\subsection{Comparing Language Models}
\label{sec:comparing}
Our NER Tagger trained on the corpus in this study was compared with four different language models: ELMo for materials synthesis (MatELMo)~\cite{doi:10.1021/acs.jcim.9b00995}\footnote{\url{https://github.com/olivettigroup/materials-synthesis-generative-models}}, BERT~\cite{Devlin2019BERTPO}\footnote{\url{https://huggingface.co/bert-base-cased}}, SciBERT~\cite{Beltagy2019SciBERT}\footnote{\url{https://huggingface.co/allenai/scibert\_scivocab\_cased}}, and PubmedBERT~\cite{pubmedbert}\footnote{\url{https://huggingface.co/microsoft/BiomedNLP-PubMedBERT-base-uncased-abstract-fulltext}} for token embedding. For SciBERT and BERT, we used transformers~\cite{wolf-etal-2020-transformers} to obtain embeddings and to connect to the NER Tagger extractor created in Flair~\cite{akbik-etal-2019-flair}, a framework for using state-of-the-art NLP models. For evaluations, the dataset was divided in a ratio of 6:2:2 for training, development, and testing, respectively. \Tref{tab:result_ner} shows the obtained results. The Mat-ELMo language model had the highest overall micro-F1 score of 0.778. SciBERT and BERT had higher F1-scores for the extraction of conductivity and activation energy, respectively. Although the present study aims to extract material names and property values, Mat-ELMo was selected for further analysis considering raw materials and temperature and time conditions as it had the highest overall micro-F1 score.

\begin{table}[t]
  \begin{center}
    \begin{tabular}{lcccc}\\
        \hline
        Model & MatELMo& PubmedBERT& SciBERT& BERT\\ \hline
        \matfinal{} & \textbf{0.613} & 0.572 & 0.595 & 0.543\\
        \matsolvent{} & \textbf{0.757} & \textbf{0.757} & 0.724 & 0.705\\
        \matstart{} & \textbf{0.754} & 0.726 & 0.688 & 0.651 \\
        \ope{} & 0.825 & \textbf{0.835} & 0.832 & 0.826 \\
        \propertyequipment & \textbf{0.819} & 0.801 & 0.798 & 0.795 \\
        \propertymanufacturer{} & \textbf{0.869} & 0.827 & 0.794 & 0.816 \\
        \propertymethod{} & 0.798 & \textbf{0.802} & 0.785 & 0.793 \\
        \propertytemp{} & 0.851 & 0.855 & \textbf{0.875} & 0.867 \\
        \propertytime{} & 0.867 & \textbf{0.892} & 0.883 & 0.887 \\
        \characteristicname{} & \textbf{0.918} & 0.912 & 0.914 & 0.914 \\
        \characteristicactivation{} & 0.593 & 0.571 & \textbf{0.654} & 0.509 \\
        \characteristicconductivity{} & 0.605 & 0.630 & 0.649 & \textbf{0.685} \\
        ALL & \textbf{0.778} & 0.767 & 0.761 & 0.749 \\ \hline \\
    \end{tabular}
    \caption{F1 scores of sequence-labeling models with different base representations on development dataset. Micro-F1 scores were calculated using all labels (ALL). The highest value is indicated in bold.}
    \label{tab:result_ner}
  \end{center}
\end{table}

\subsection{Tuning Hyperparameters}
We performed hyperparameter tuning for the NER model employing Mat-ELMo, which showed the best extraction performance as described in \Sref{sec:comparing}. We used optuna~\cite{optuna_2019}, a sophisticated optimization tool, for exploring the parameter space. \Tref{tab:tuning} summarizes the evaluated hyperparameter space and the best parameters used for the final evaluation on the test dataset. After obtaining the optimal hyperparameter values, the model was trained again to evaluate its final performance. As a result shown in \Tref{tab:finalresult}, after tuning the hyperparameters, the micro-F1 score was improved from 77.8\% to 78.1\%.

\begin{figure}
    \begin{minipage}[c]{0.45\textwidth}
	    \begin{center}
            \begin{tabular}{lc}\hline
                Label & F1-score \\ \hline
                \matfinal{} & 0.625 \\
                \matsolvent{} & 0.771 \\
                \matstart{} & 0.771 \\
                \ope{} & 0.827 \\
                \propertyequipment{} & 0.827 \\
                \propertymanufacturer{} & 0.870 \\
                \propertymethod{} & 0.813 \\
                \propertytemp{} & 0.857 \\
                \propertytime{} & 0.869 \\
                \characteristicname{} & 0.917 \\
                \characteristicactivation{} & 0.593 \\
                \characteristicconductivity{} & 0.637 \\
                ALL & 0.781 \\ \hline \\
            \end{tabular}
		\subcaption{Final result. ALL is the micro-F1 score.}
		\label{tab:finalresult}
		\end{center}
	\end{minipage}%
    \begin{minipage}[c]{0.45\textwidth}
	    \begin{center}
            \begin{tabular}{lcc}\hline
                Parameter & Range & Best \\ \hline
                Learning rate & [0.05, 0.3] & 0.15\\
                Dropout & [0.3, 0.6] & 0.3 \\
                Locked dropout & [0.3, 0.6] & 0.4 \\
                Word dropout & [0.05, 0.15] & 0.1 \\
                Hidden size & [32, 256] & 256 \\
                RNN layers & [1, 3] & 2 \\
                Weight decay & [0.0001, 0.0005] & 0.0005 \\ \hline \\
            \end{tabular}
		\subcaption{Hyperparameters}
		\label{tab:tuning}
		\end{center}
	\end{minipage}%
	\caption{Micro-F1 score after hyperparameter tuning.}
\end{figure}

\subsection{Evaluation of Extracted NE Result}
\begin{figure}
	\begin{minipage}[c]{0.45\textwidth}
	    \begin{center}
		\begin{tabular}{lc}\hline
		    Label & Recall \\ \hline
		    \matfinal{} & 0.873 \\
		    \matsolvent{} & 0.956 \\
		    \matstart{} & 0.751  \\
		    \ope{} & 0.997 \\
		    \propertyequipment{} & 0.990 \\
		    \propertymanufacturer{} & 1.000 \\
		    \propertymethod{} & 0.987 \\
		    \propertytemp{} & 0.924 \\
		    \propertytime{} & 0.964 \\
		    \characteristicname{} & 0.994 \\
		    \characteristicactivation{} & 1.000 \\
		    \characteristicconductivity{} & 0.926 \\
		    ALL & 0.920 \\
		    \hline \\
		\end{tabular}
		\subcaption{Recall of each label. ALL is the macro recall score.}
		\label{tab:post_eval_recall}
		\end{center}
	\end{minipage}%
	\begin{minipage}[h]{0.45\textwidth}
		\begin{center}
			\includegraphics[clip, width=1.1\textwidth]{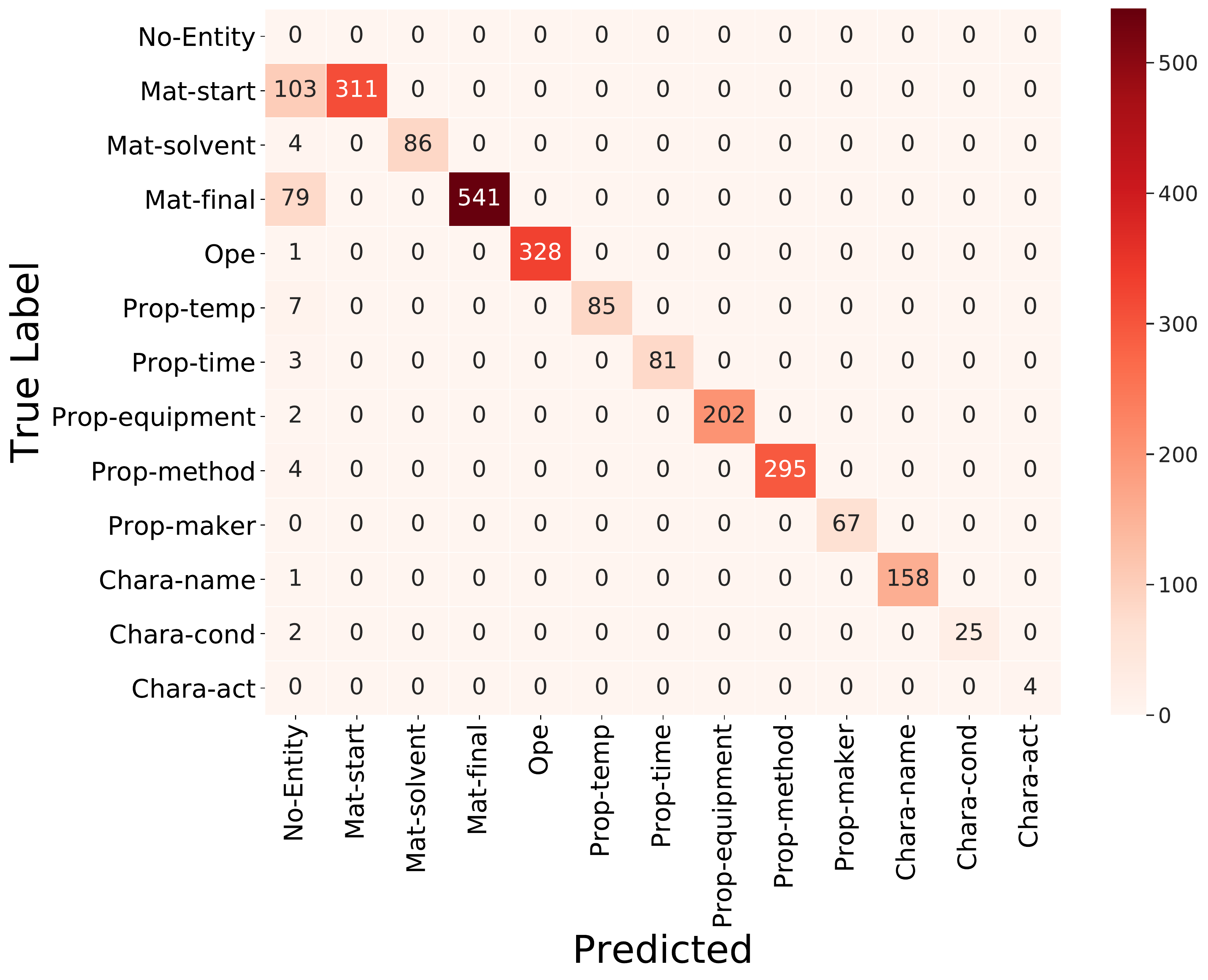}
		\subcaption{Post-evaluation results}
		\label{fig:post_eval_confusion}
		\end{center}
	\end{minipage}
	\caption{Post-evaluation}
	\label{fig:post_eval}
\end{figure}

To verify the extraction performance of the tuned NER model, 100 paragraphs from 10 papers were labeled by NER. These paragraphs cover all sections that can be extracted from a paper. The labels were checked by the same annotator who labeled the corpus to correct any extraction errors or omissions. These validations were performed to evaluate the machine extraction performance when the machine's labeling was considered the correct answer, and the recall metric was used to evaluate the extraction performance of correctly extracted entities among the entities labeled by the NER model. \Tref{tab:post_eval_recall} shows the agreement between our NER model and the human annotators for each label, and \Tref{fig:post_eval_confusion} shows the confusion matrix.

\section{Research Trends Analysis}
In this section, we analyze the NLP outputs to understand the trend of materials by year. \Tref{tab:stat_token} summarizes several key statistics of the NLP outputs, such as the number of papers, entities, and distribution of converted values. Further, \Tref{tab:stat_country} shows a country-by-country tabulation. Only abstracts extracted from papers were used for this tabulation, and the first author's country was counted.

\begin{figure}
\begin{center}
	\begin{minipage}[c]{0.45\textwidth}
	    \begin{center}
    		\begin{tabular}{lr}\hline
                Item & Count  \\ \hline
                Papers & 12,895 \\
                Paragraphs & 57,783 \\ \hline
                \matfinal{} & 919,645 \\
                \matsolvent{} & 63,437 \\
                \matstart{} & 406,387  \\
        	    \ope{} & 277,418 \\
        	    \propertyequipment{} & 175,299 \\
        	    \propertymanufacturer{} & 55,018 \\
        	    \propertymethod{} & 454,945 \\
        	    \propertytemp{} & 75,004 \\
        	    \propertytime{} & 77,889 \\
        	    \characteristicname{} & 67,530 \\
        	    \characteristicactivation{} & 14,596 \\
        	    \characteristicconductivity{} & 31,005 \\ \hline \\
            \end{tabular}
    		\subcaption{Extracted data statistics.}
    		\label{tab:stat_token}
		\end{center}
	\end{minipage}%
	\begin{minipage}[h]{0.45\textwidth}
		\begin{center}
    		\begin{tabular}{lr}\hline
                Country & Count  \\ \hline
                China & 7593 \\
                United States & 947 \\
                Korea & 884 \\
                Japan & 384 \\
                India & 365 \\
                UK & 342 \\
                Germany & 305 \\
                Australia & 289 \\
                Spain & 207 \\
                Singapore & 204 \\
                Taiwan & 150 \\
                France & 133 \\
                Canada & 113 \\
                Sweden & 99 \\ \hline \\
            \end{tabular}
    		\subcaption{Top 14 countries.}
    		\label{tab:stat_country}
		\end{center}
	\end{minipage}
	\caption{Base statistics}
\end{center}
\end{figure}

Next, we aggregated the extracted final materials by year for a quick analysis of the trends by year, as shown in~\Tref{tab:year_ranking}. Only the paragraph section name ``Abstract'' was used in the extracted papers to prevent double-counting of papers. This result shows that the trend of frequently used materials differs by year. For example, ``TiO$_2$'' is ranked fourth in 2016--2017; however, it is ranked first in 2018--2019.

\begin{table}[t]
  \begin{center}
    \begin{tabular}{cc}\\
        \hline
        2016 -- 2017 & 2018 -- 2019\\ \hline
        reduced graphene oxide(21) & TiO$_2$(16) \\
        CH$_3$NH$_3$PbI$_3$(20) & reduced graphene oxide(15) \\
        graphene(18) & graphene(13) \\
        TiO$_2$(17) & CH$_3$NH$_3$PbI$_3$(10) \\
        carbon(17) & SnO$_2$(10) \\
        graphene oxide(10) & carbon(9) \\
        ZnO(9) & MAPbI$_3$(9) \\
        PEDOT:PSS(8) & MoS$_2$(8) \\
        activated carbon(7) & covalent organic frameworks(8)\\
        BiVO$_4$(6) & MOFs(8)\\
        \hline \\
    \end{tabular}
    \caption{Reported material final aggregation by year. Number in the brackets means number of extracted phrases.}
    \label{tab:year_ranking}
  \end{center}
\end{table}

\begin{figure}[t]
  \centering
  \begin{minipage}[b]{0.48\textwidth}
    \centering
    \includegraphics[keepaspectratio, scale=0.16]{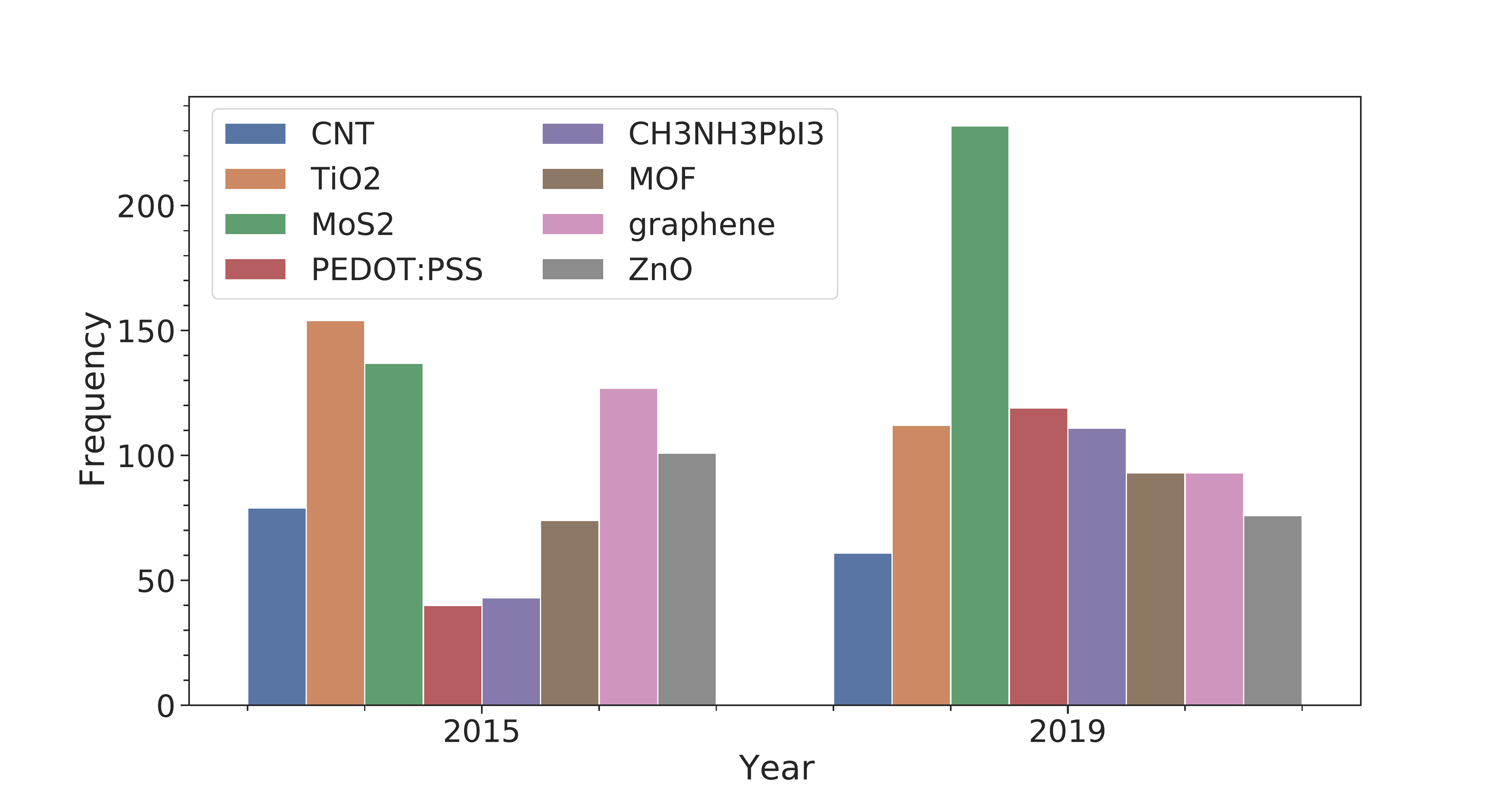}
    \subcaption{Reported materials in China}
    \label{fig:bar_china}
  \end{minipage}
  \begin{minipage}[b]{0.48\textwidth}
    \centering
    \includegraphics[keepaspectratio, scale=0.16]{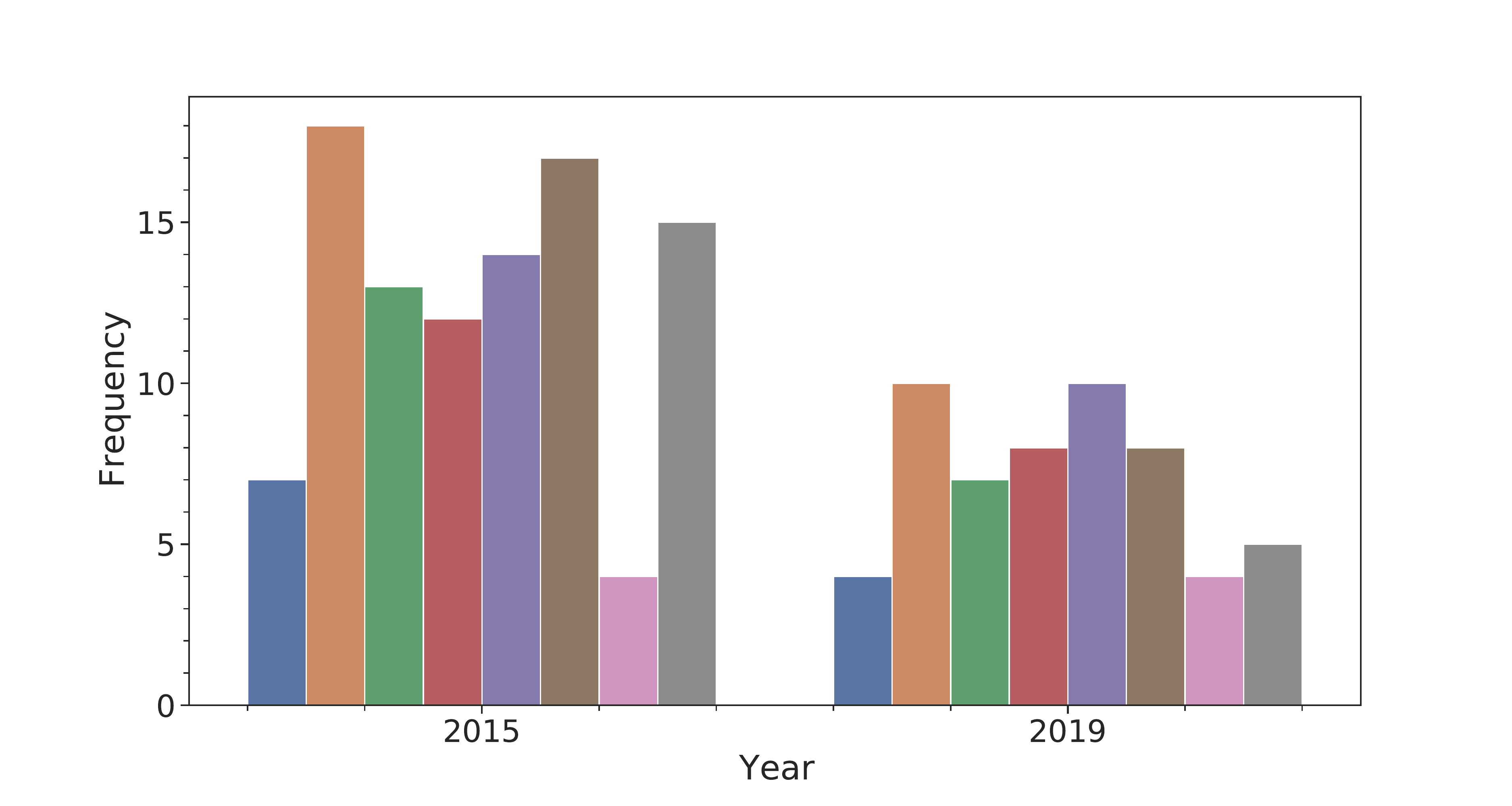}
    \subcaption{Reported materials in United States}
    \label{fig:bar_USA}
  \end{minipage}
  \caption{The year of transition of material by country and by year.}
  \label{fig:bar_country}
\end{figure}

For the following analysis, we manually selected the following final materials from ~\Tref{tab:year_ranking} for a query search using the word2vec model~\cite{Mikolov2013DistributedRO} to efficiently screen the many other materials obtained from extracted outputs: ``CH$_3$NH$_3$PbI$_3$,'' ``PEDOT:PSS,'' ``TiO$_2$,'' ``graphene,'' ``ZnO,'' ``MoS$_2$,'' ``MOF,'' and ``CNT.'' This model was trained using the same 12,895 papers that were input to the NLP pipeline. We then applied the trained word2vec model to the extracted final material, the query with the highest cosine similarity was adopted as the type. The final material classified by type is used for trend analysis by country, as shown in~\Fref{fig:bar_country}. This figure shows the features of developed materials by country and the change in the number of reported materials by year. Consider the comparison of China~\tref{fig:bar_china} and United States~\tref{fig:bar_USA}: in China, the number of papers on ``MoS$_2$'' has been increasing in recent years, whereas in the United States, the number of papers has been decreasing. Further, the most frequently reported material in China in 2015 was ``TiO$_2$,'' whereas that in 2019 was ``MoS$_2$,'' indicating the shifting trend in materials science research.

\begin{figure}[t]
  \centering
  \begin{minipage}[b]{0.47\textwidth}
    \centering
    \includegraphics[keepaspectratio, scale=0.2]{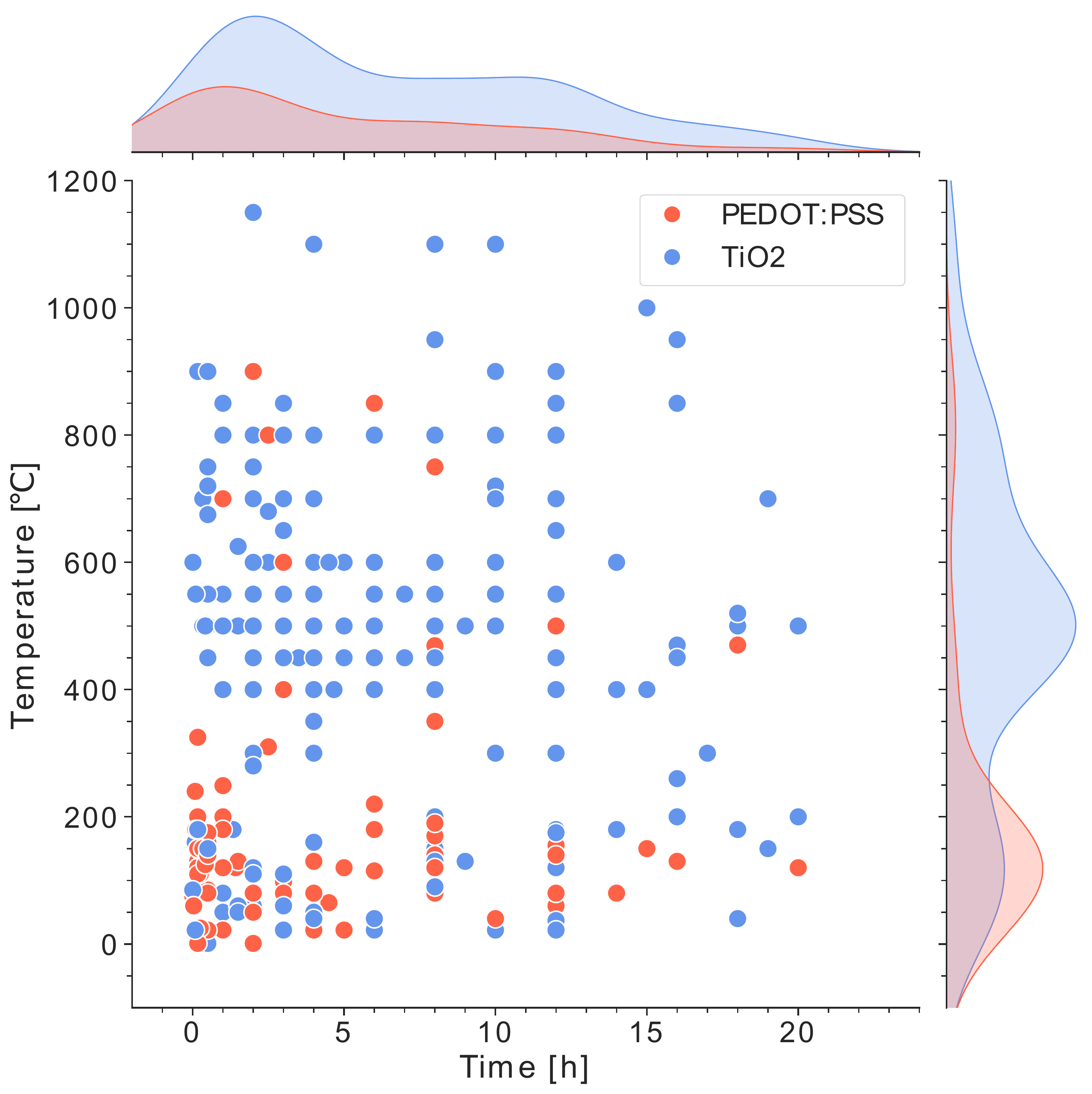}
    \subcaption{Temperature-by-time in 2015}
    \label{fig:condition2015}
  \end{minipage}
  \begin{minipage}[b]{0.47\textwidth}
    \centering
    \includegraphics[keepaspectratio, scale=0.2]{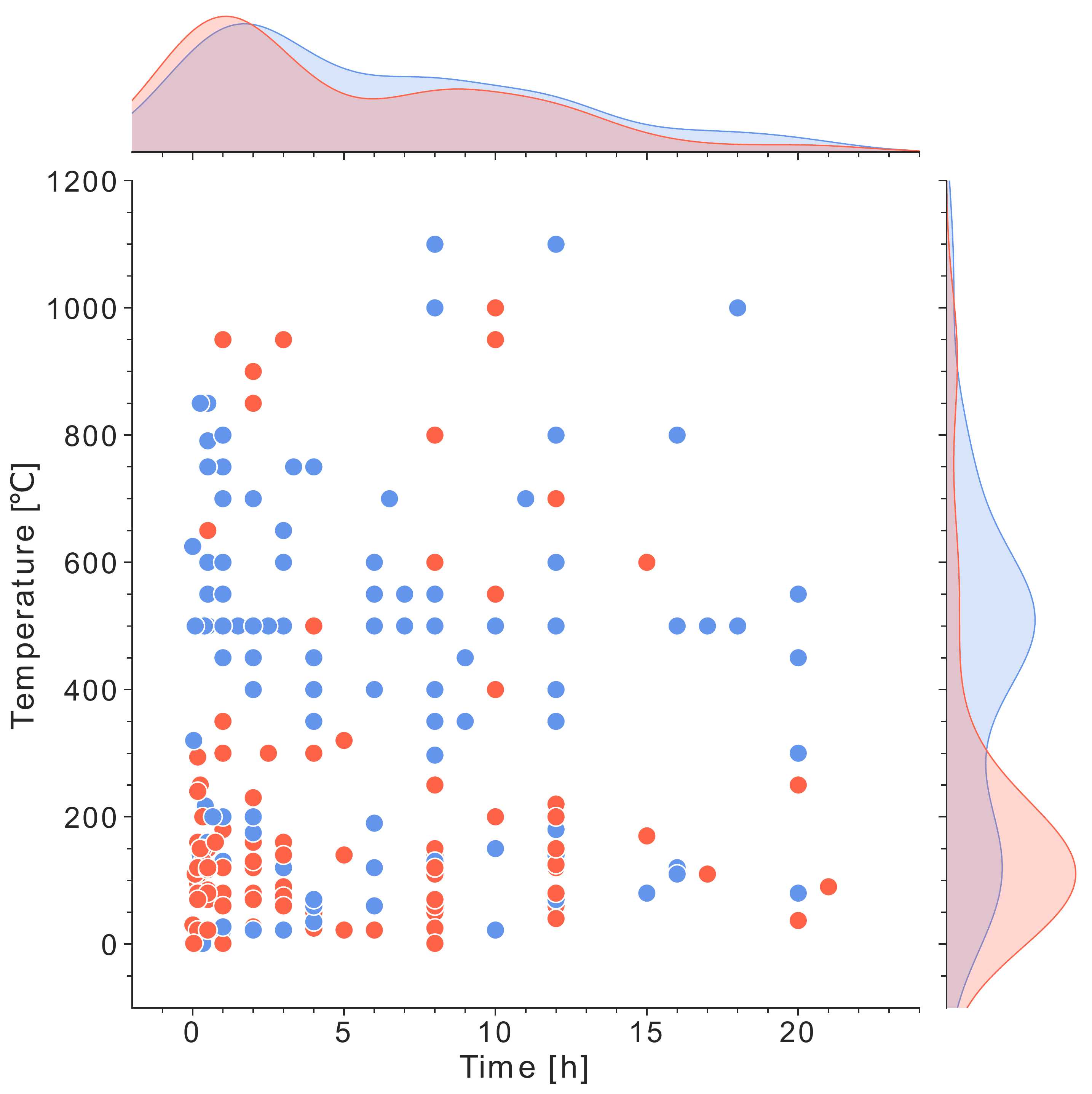}
    \subcaption{Temperature-by-time in 2018}
    \label{fig:condition2018}
  \end{minipage}
  \caption{Condition-by-year.}
  \label{fig:condition-by-year}
\end{figure}

We also visualized the condition-by-year for temperature and time, as shown in \Fref{fig:condition-by-year}. If multiple properties are extracted from a single paragraph, we select the property with the highest characteristic value. These results show that the trends of temperature and time when synthesizing ``PEDOT:PSS'' and ``TiO$_2$'' vary by year. In particular, the processing temperature of ``PEDOT:PSS'' shifted below 200$^{\circ}$C. Further, the processing times of ``PEDOT:PSS'' and ``TiO$_2$'' differed in 2015, and the processing temperatures were similar in 2019. This indicates that there are similarities in the synthesis methods of ``PEDOT:PSS'' and ``TiO$_2$.'' Other visualization examples are introduced in the supplementary information.We have received comments from one material researcher that the results reported in this study are useful when investigating competitors and when designing material synthesis processes outside the laboratory.

\section{Conclusion}
This study proposed an NLP-based approach to analyze the trend in materials research for development in materials science. 
We developed an NLP system with the BiLSTM-CRF model that was trained using manually labeled literature for extracting material properties from scientific literature. We conducted experiments to verify the effectiveness of the proposed NER method in the field of materials science.

The present study has two limitations: (1) linking multiple materials and property values when they are written in a single document, and (2) extracting characteristic values from nontextual components such as charts, diagrams, and tables that provide key information in many scientific documents. We aim to overcome these limitations through our ongoing studies.

In future work, we will aim to predict characteristic values such as conductivity and materials research trends from previous scientific literature to achieve computational materials synthesis.

\bibliographystyle{splncs04}
\bibliography{main}

\end{document}


\maketitle

In this document, we present a visualization example of the output obtained from our NLP pipeline.

\begin{figure}[t]
  \begin{minipage}[b]{0.45\hsize}
    \centering
    \includegraphics[keepaspectratio, scale=0.3]{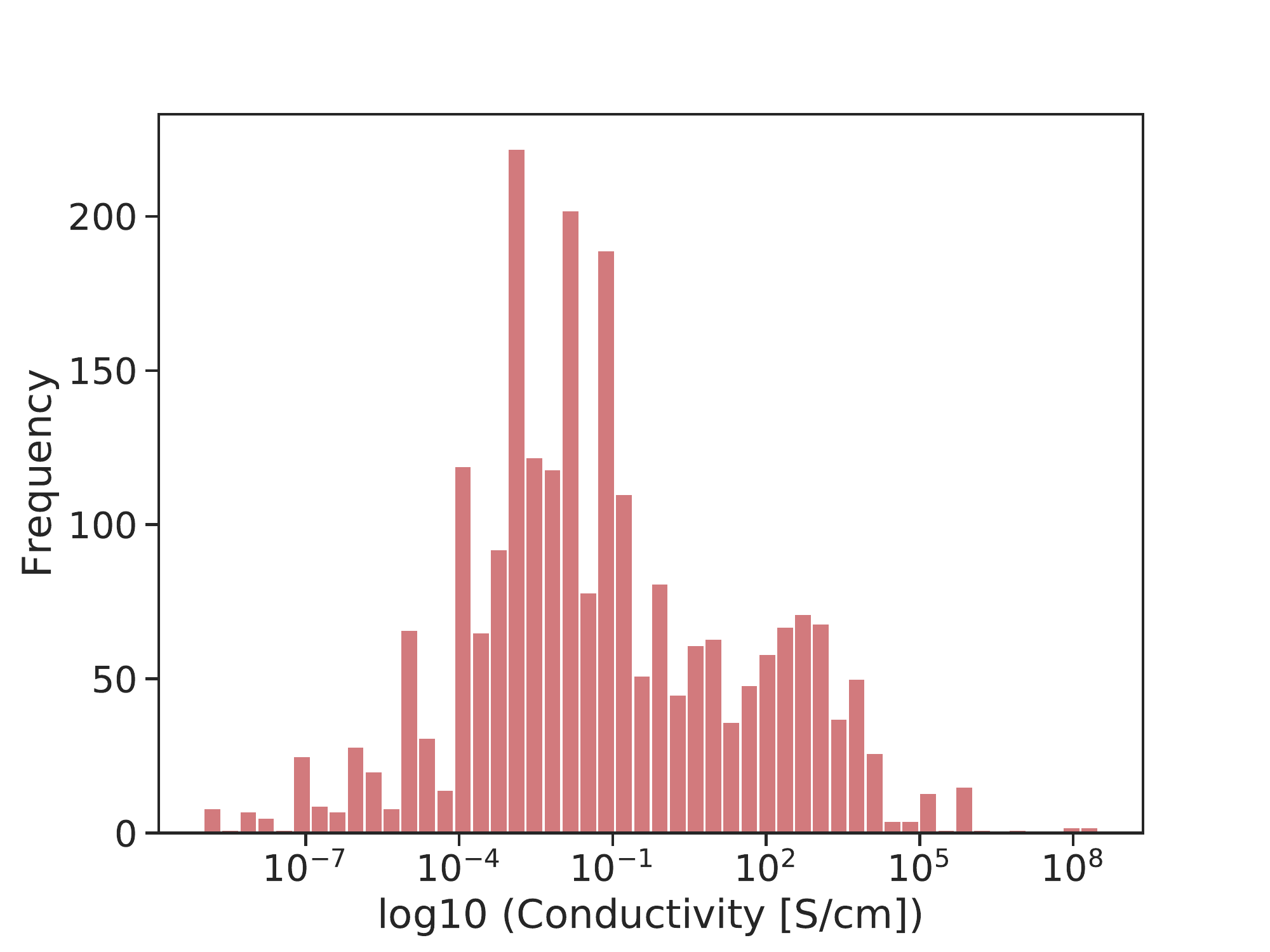}
    \subcaption{Conductivity}
    \label{fig:conductivity}
  \end{minipage}
  \begin{minipage}[b]{0.45\hsize}
    \centering
    \includegraphics[keepaspectratio, scale=0.3]{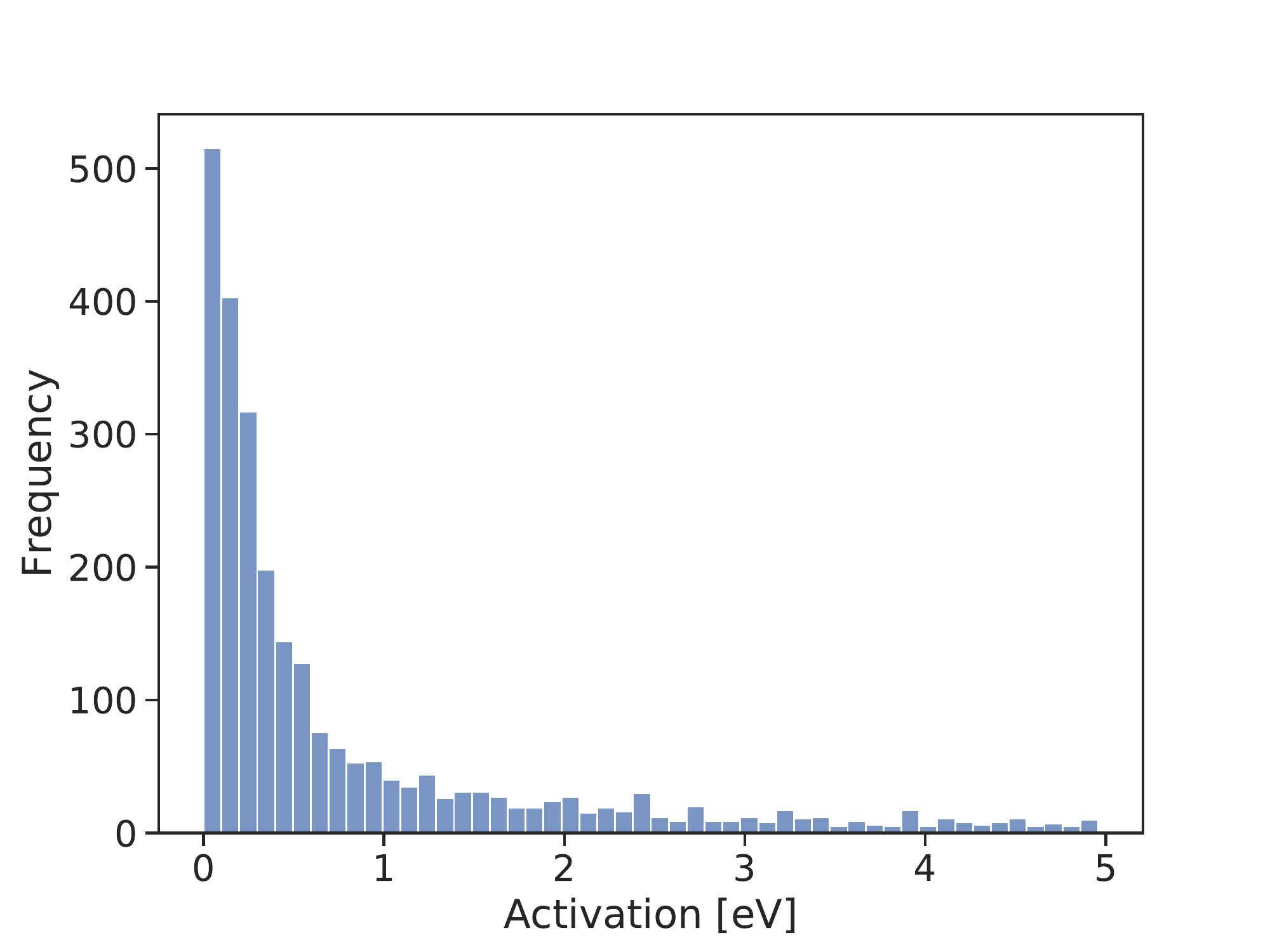}
    \subcaption{Activation energy}
    \label{fig:activation}
  \end{minipage}\\
  \begin{minipage}[b]{0.45\hsize}
    \centering
    \includegraphics[keepaspectratio, scale=0.3]{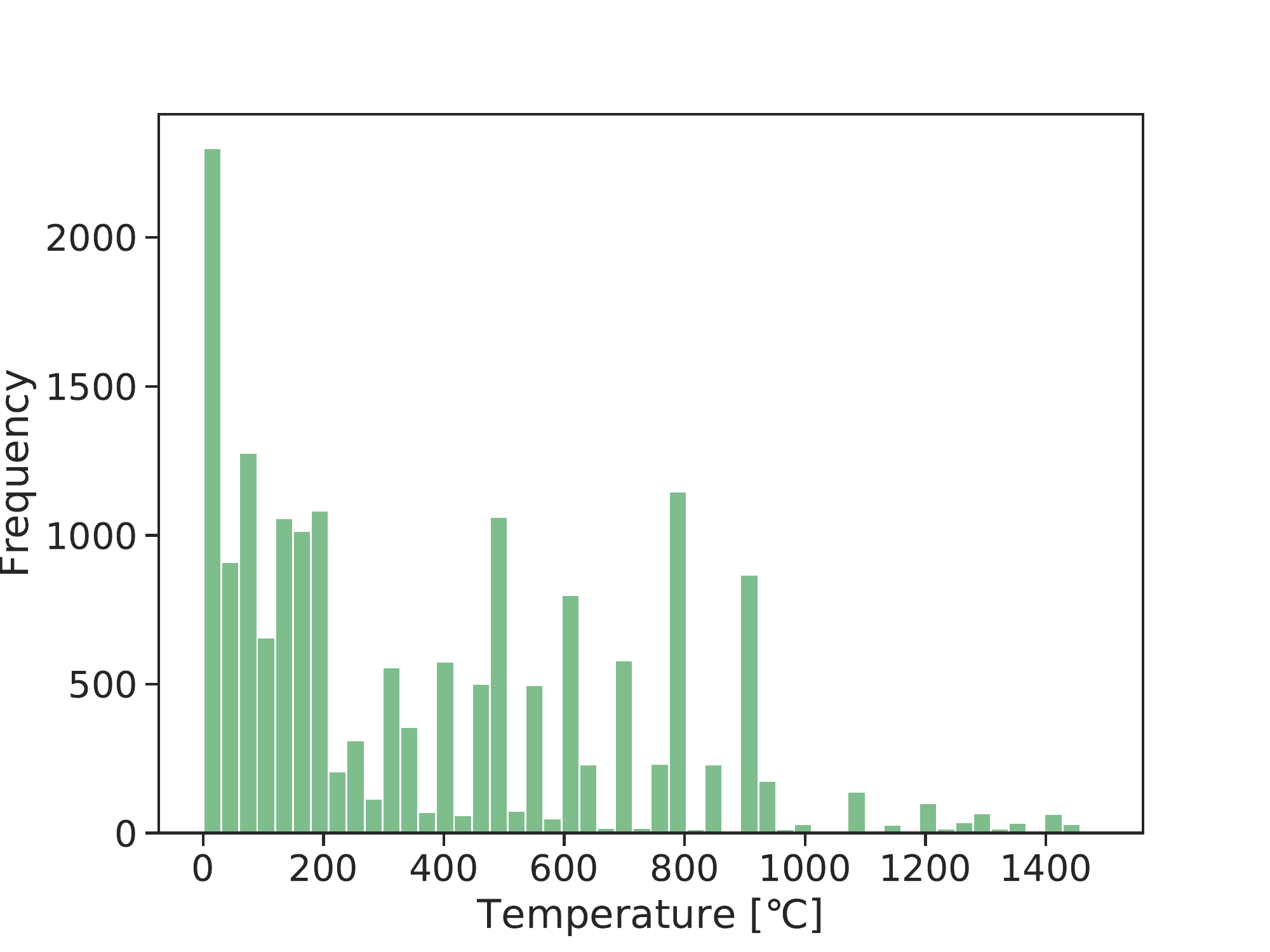}
    \subcaption{Temperature}
    \label{fig:temp}
  \end{minipage}
  \begin{minipage}[b]{0.45\hsize}
    \centering
    \includegraphics[keepaspectratio, scale=0.3]{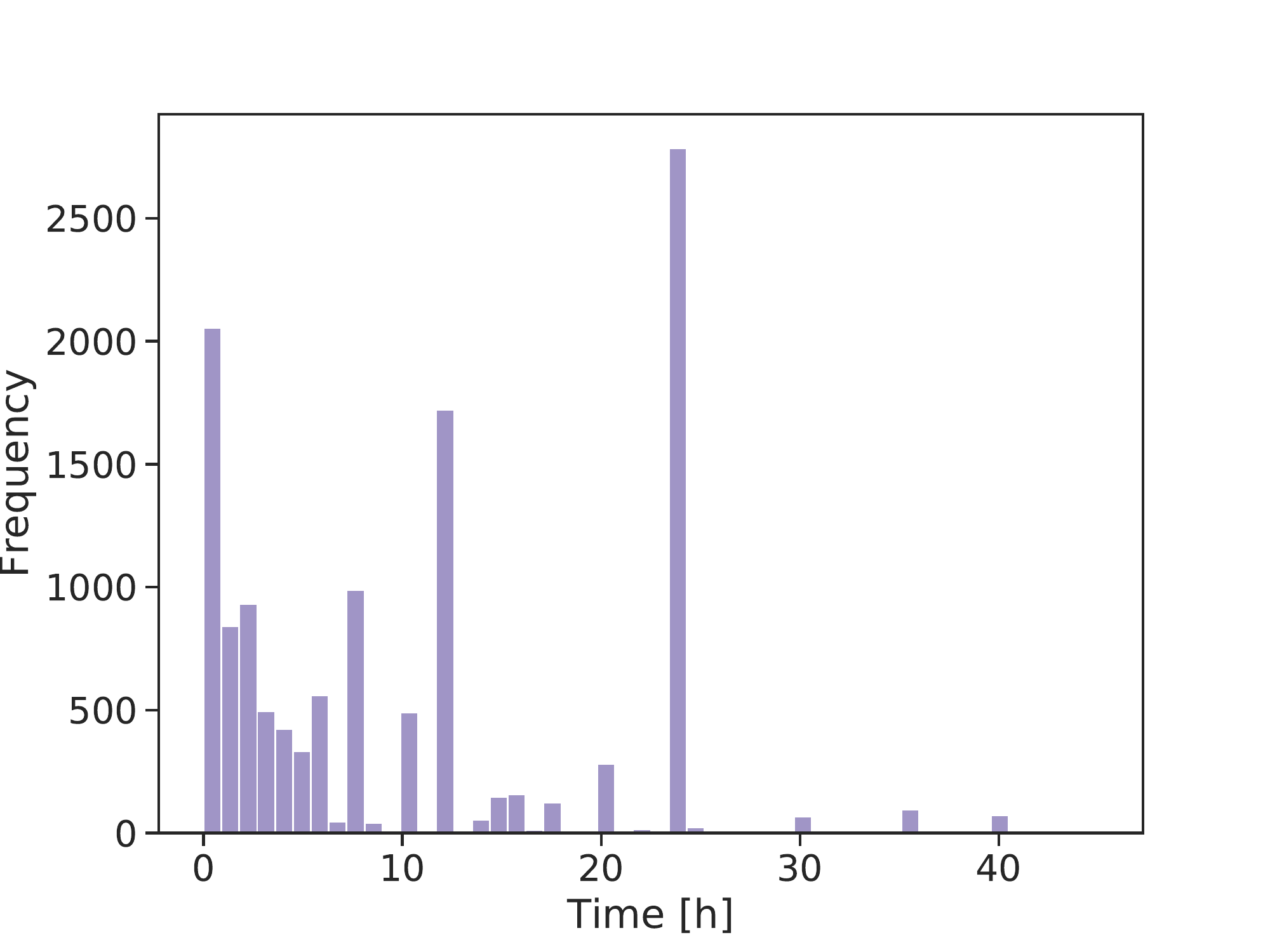}
    \subcaption{Time}
    \label{fig:time}
  \end{minipage}
  \caption{Distributions of four properties in our automatically generated datasets. X-axis shows each property per paragraph. Y-axis shows fraction of paragraphs with properties: (a) conductivity, (b) activation energy, (c) temperature, and (d) time.}
  \label{fig:value_dist}
\end{figure}

\textbf{Distributions of properties}: \Fref{fig:value_dist} is a visualization of the distribution of the four properties in our automatically generated datasets. The conductivity peaks around 10$^{-3}$ S/cm and 10$^2$ S/cm, while the activation energy is concentrated below 0.3 eV. In addition, the temperature is sparsely distributed below 1000 $^{\circ}$C, while the most common time is 24 hours.

\begin{figure}[t]
  \begin{minipage}[b]{\hsize}
    \centering
    \includegraphics[keepaspectratio, scale=0.23]{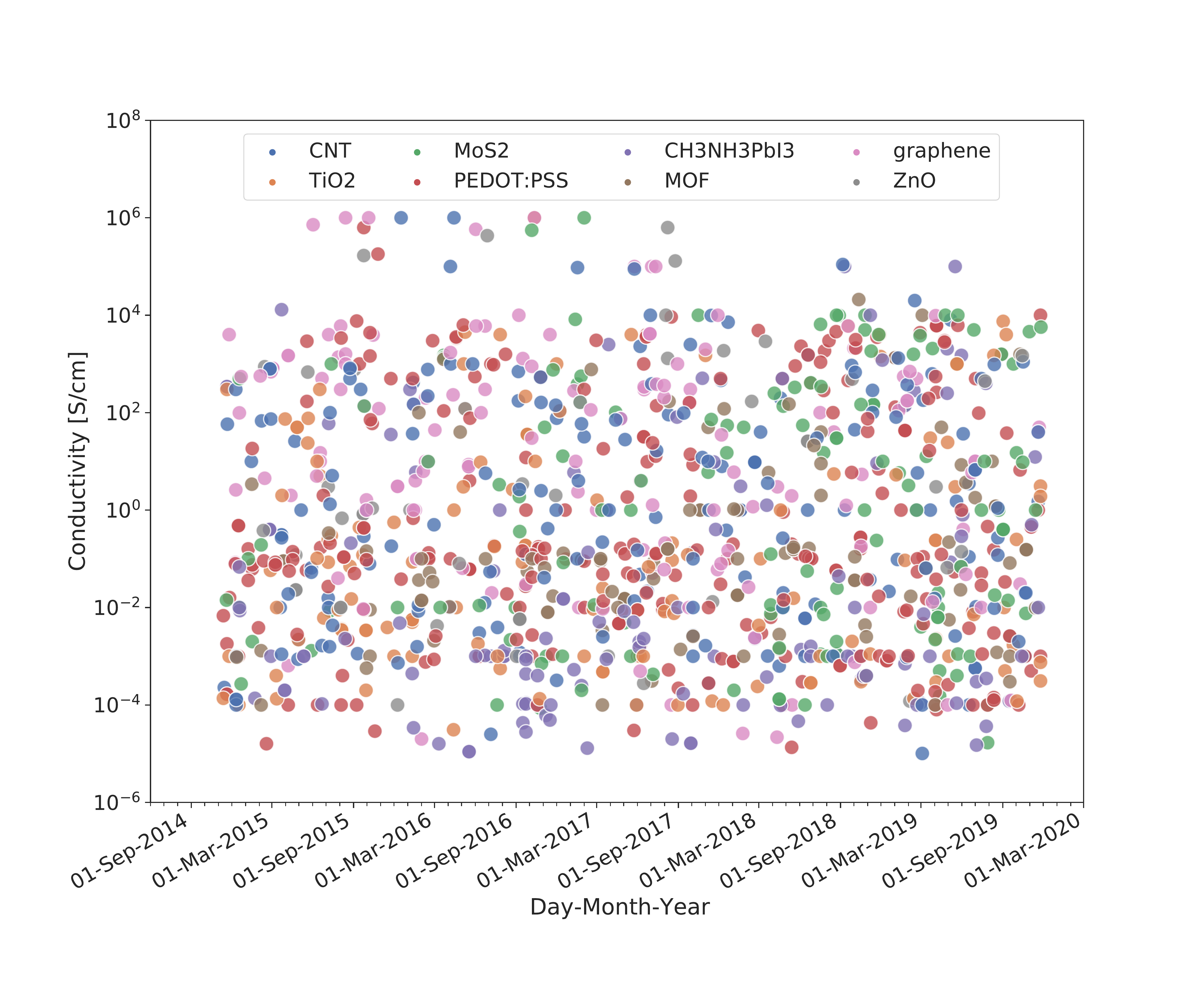}
    \subcaption{Conductivity-by-year.}
    \label{fig:conductivity-by-year}
  \end{minipage}\\
  \begin{minipage}[b]{\hsize}
    \centering
    \includegraphics[keepaspectratio, scale=0.23]{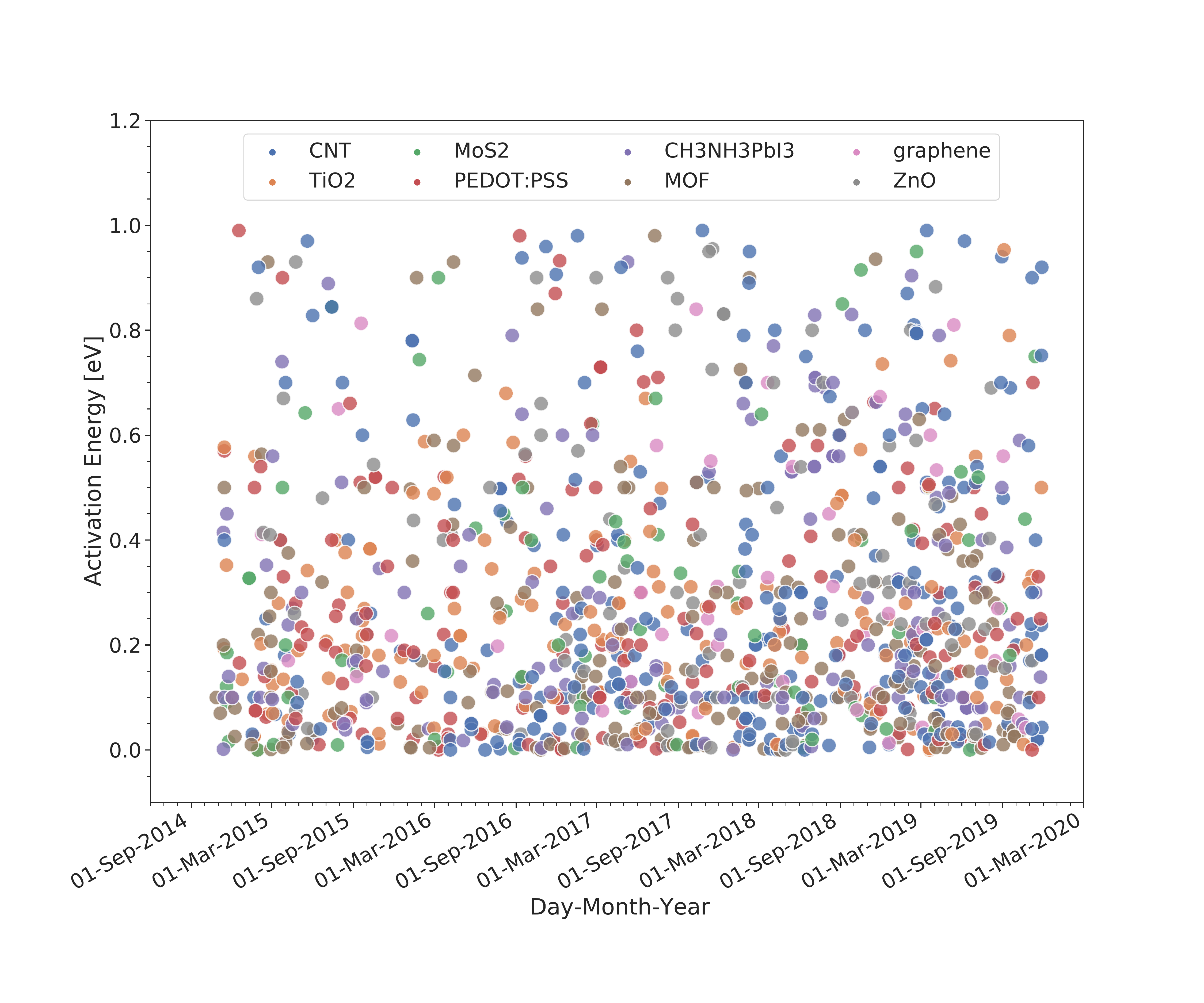}
    \subcaption{Activation-by-year.}
    \label{fig:activation-by-year}
  \end{minipage}
  \caption{Value-by-year.}
  \label{fig:value-by-year}
\end{figure}

\textbf{Value-by-year}: \Fref{fig:conductivity-by-year} and \Fref{fig:activation-by-year} are scatter plots of the extracted conductivity and activation energy by year, respectively.  The pixels are colored based on categorical materials identified by word2vec. From this figure, it can be seen that many ``MoS$_2$'' are reported in the area with high conductivity from May 2018 onwards.

\begin{figure}[t]
  \begin{minipage}[b]{\hsize}
    \centering
    \includegraphics[keepaspectratio, scale=0.23]{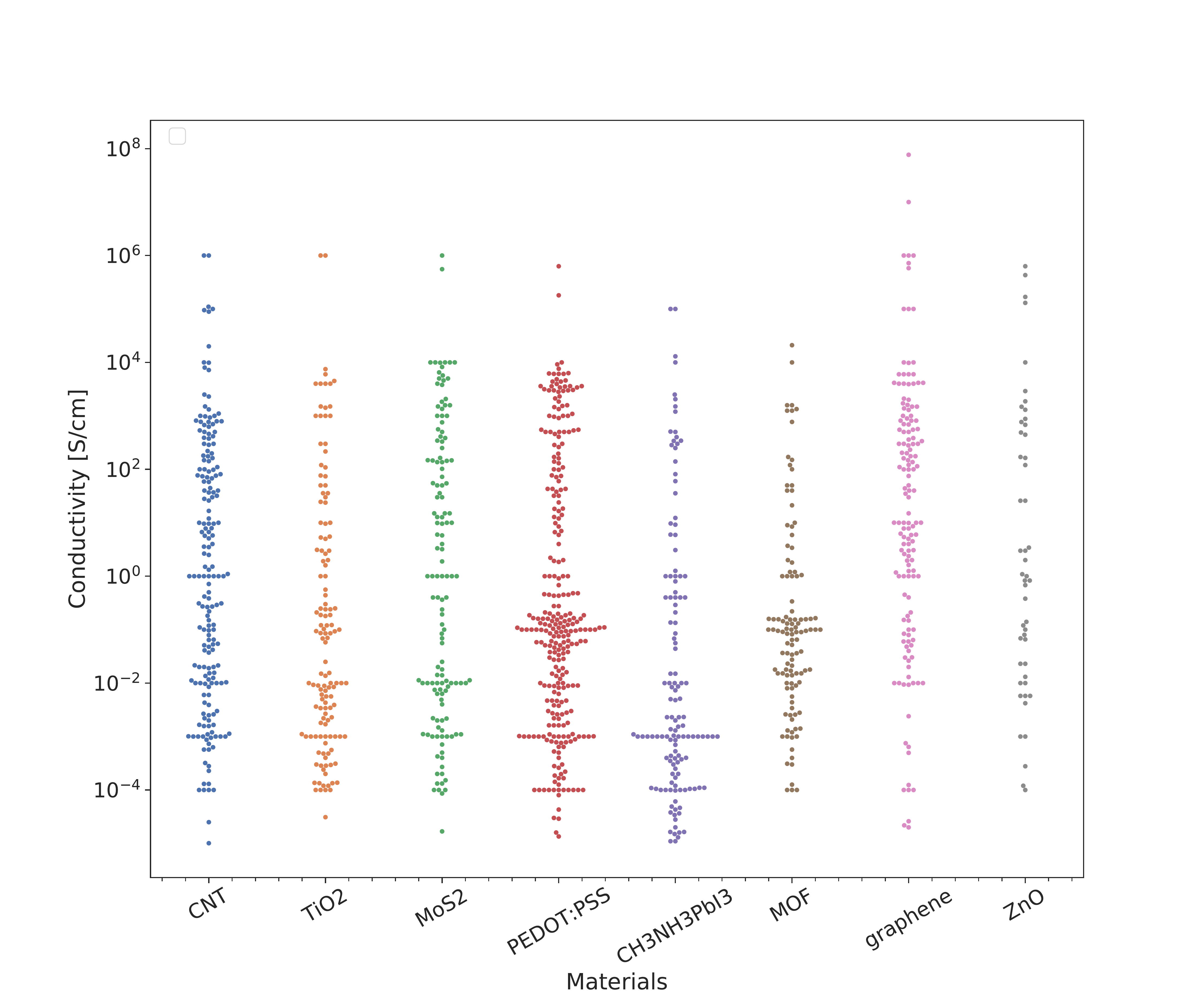}
    \subcaption{Conductivity-by-materials.}
    \label{fig:conductivity-by-material}
  \end{minipage}\\
  \begin{minipage}[b]{\hsize}
    \centering
    \includegraphics[keepaspectratio, scale=0.23]{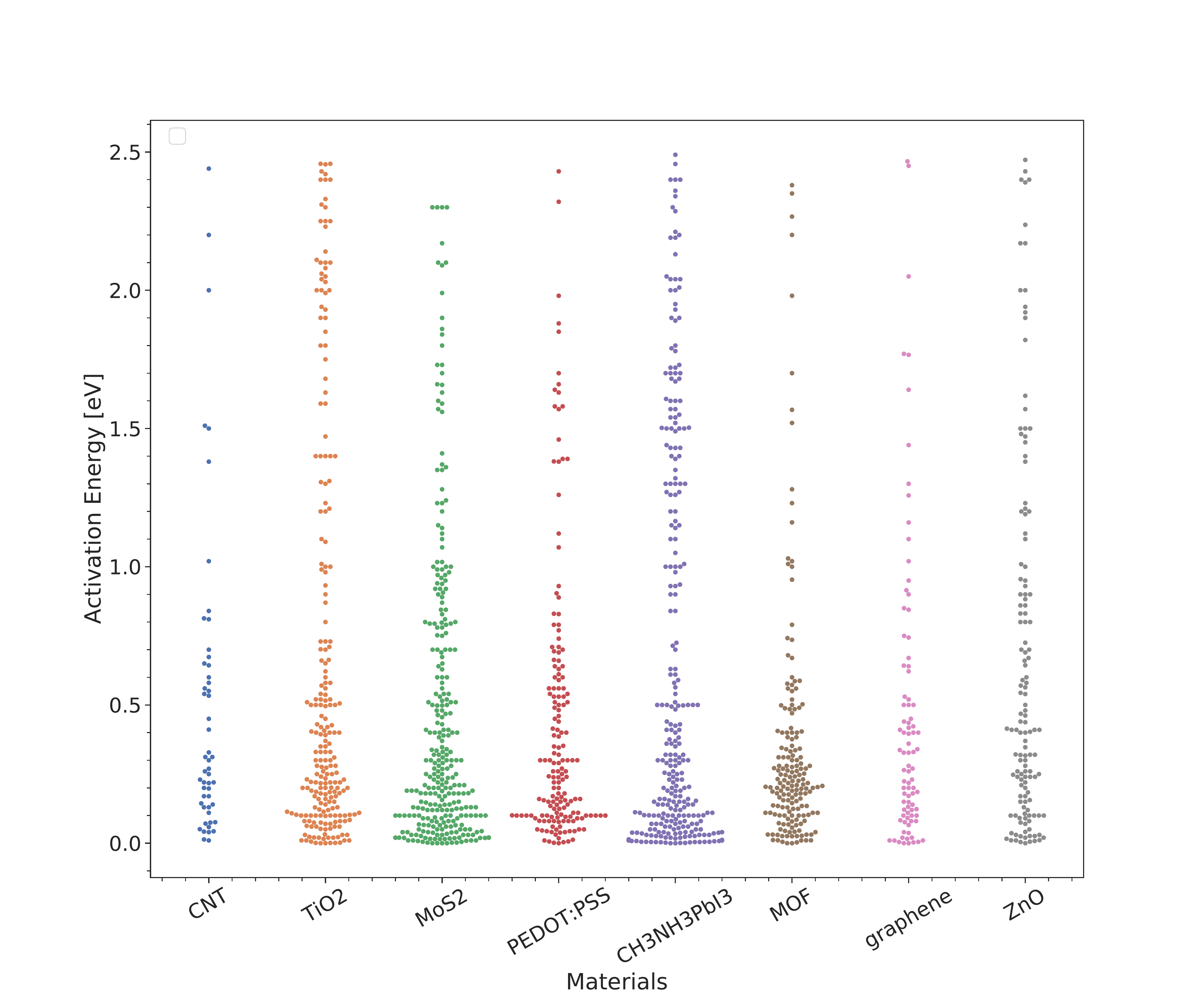}
    \subcaption{Activation-by-materials.}
    \label{fig:activation-by-material}
  \end{minipage}
  \caption{Value-by-materials.}
  \label{fig:value-by-mat}
\end{figure}

\textbf{Value-by-materials}: \Fref{fig:conductivity-by-material} and \Fref{fig:activation-by-material} are swarm plots of the extracted conductivity and activation energy by year, respectively. It was found that the distribution of the characteristic values was different for each material, and the activation energy was concentrated and distributed below 0.5 eV.

\begin{figure}[t]
 \centering
 \includegraphics[scale=0.2]{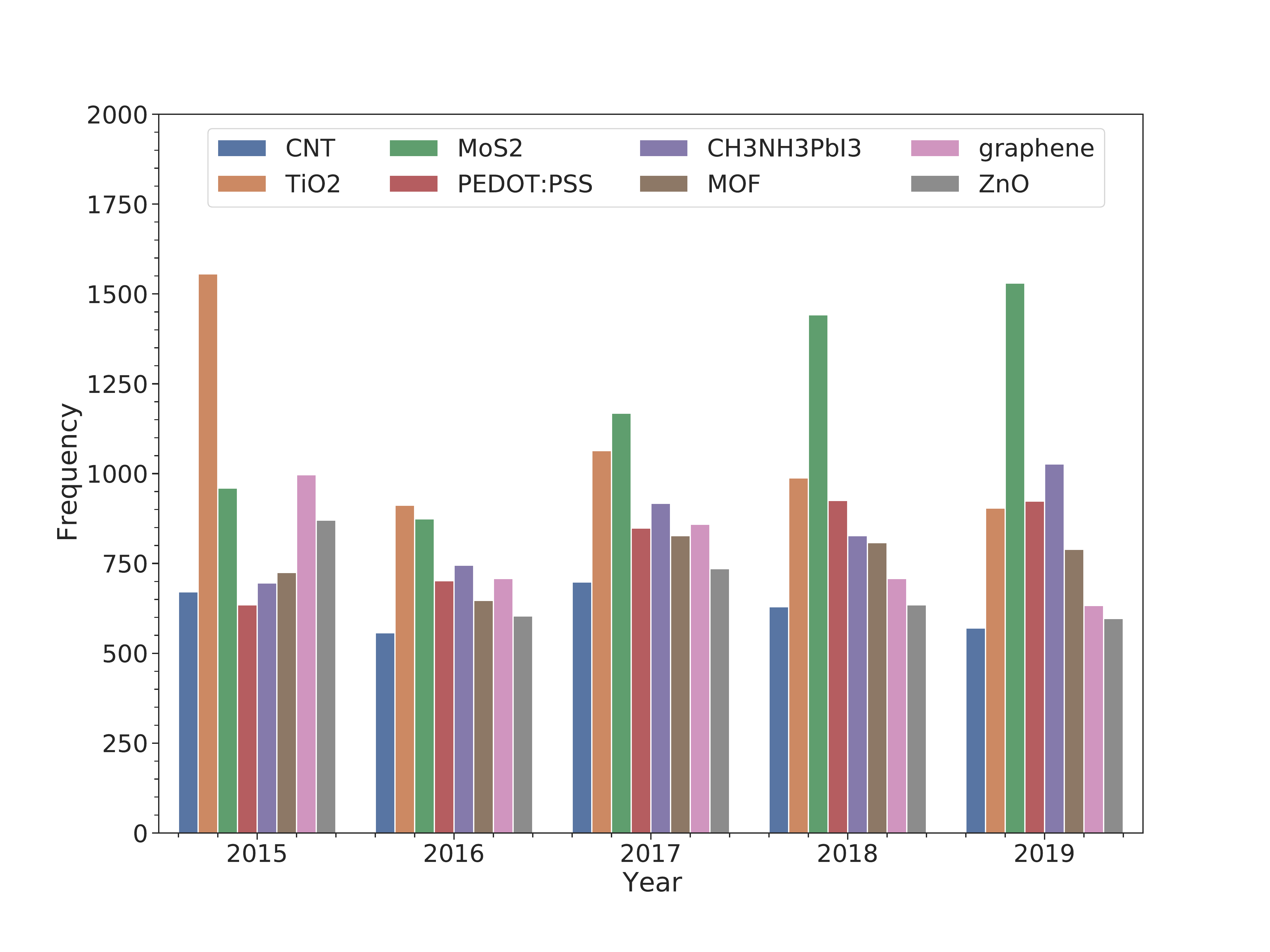}
 \caption{Frequency of materials by year.}
 \label{fig:freqmat_by_year}
\end{figure}

\textbf{Materials-by-year}: \Fref{fig:freqmat_by_year} shows frequency of materials by year. In this figure, we found that the number of ``MoS$_2$'' is increasing every year, while the number of ``TiO$_2$ is decreasing after peaking in 2015. 

\begin{figure}[t]
 \centering
 \includegraphics[width=\linewidth]{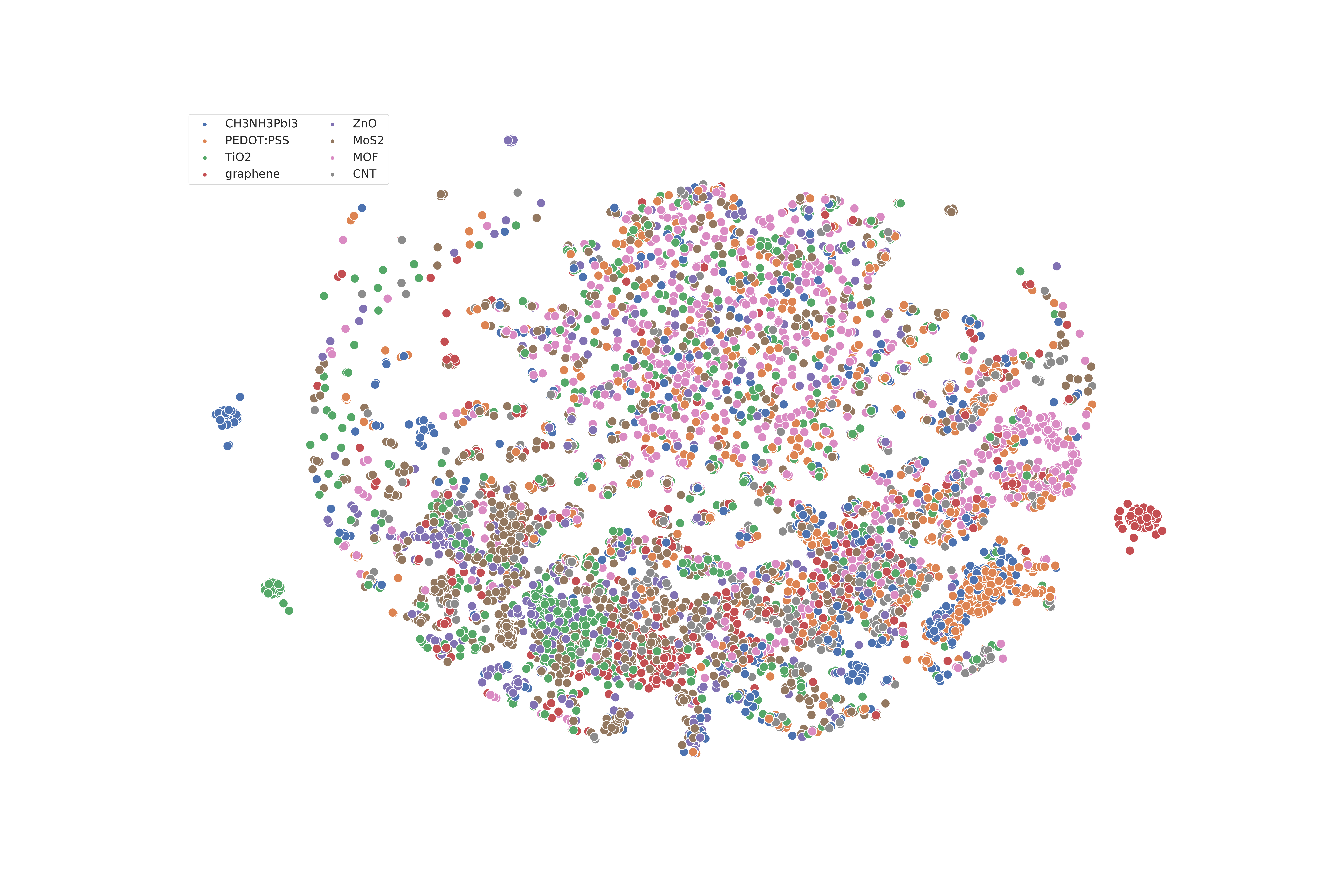}
 \caption{Clusters of vectorized \matfinal{} colored by type classified by word2vec.}
 \label{fig:tsne}
\end{figure}

\textbf{The clusters of vectorized \matfinal{}}: \Fref{fig:tsne} is a graphical representation of vectorized \matfinal{} using word2vec and t-SNE. The pixels are colored based on categorical materials identified by word2vec. We observe that ``graphene'', ``CH$_3$NH$_3$PbI$_3$'' and ``TiO$_2$'' formed small clusters, while ``CH$_3$NH$_3$PbI$_3$'' and ``PEDOT:PSS'' tended to be vectored to the same position.

\begin{figure}[t]
  \centering
  \begin{minipage}[b]{0.45\hsize}
    \centering
    \includegraphics[keepaspectratio, scale=0.15]{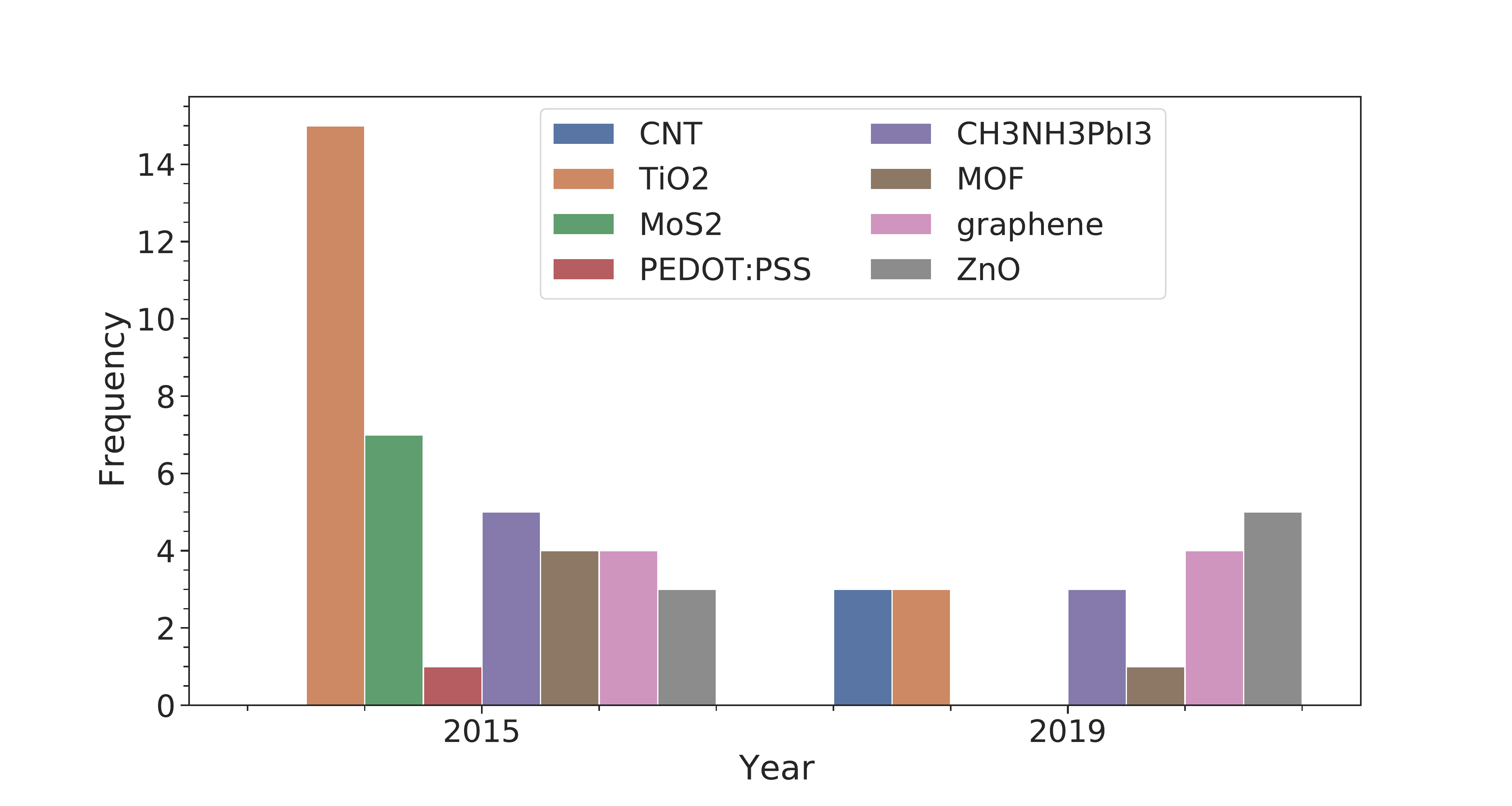}
    \subcaption{Reported materials in Japan}
    \label{fig:bar_Japan}
  \end{minipage}
  \begin{minipage}[b]{0.45\hsize}
    \centering
    \includegraphics[keepaspectratio, scale=0.15]{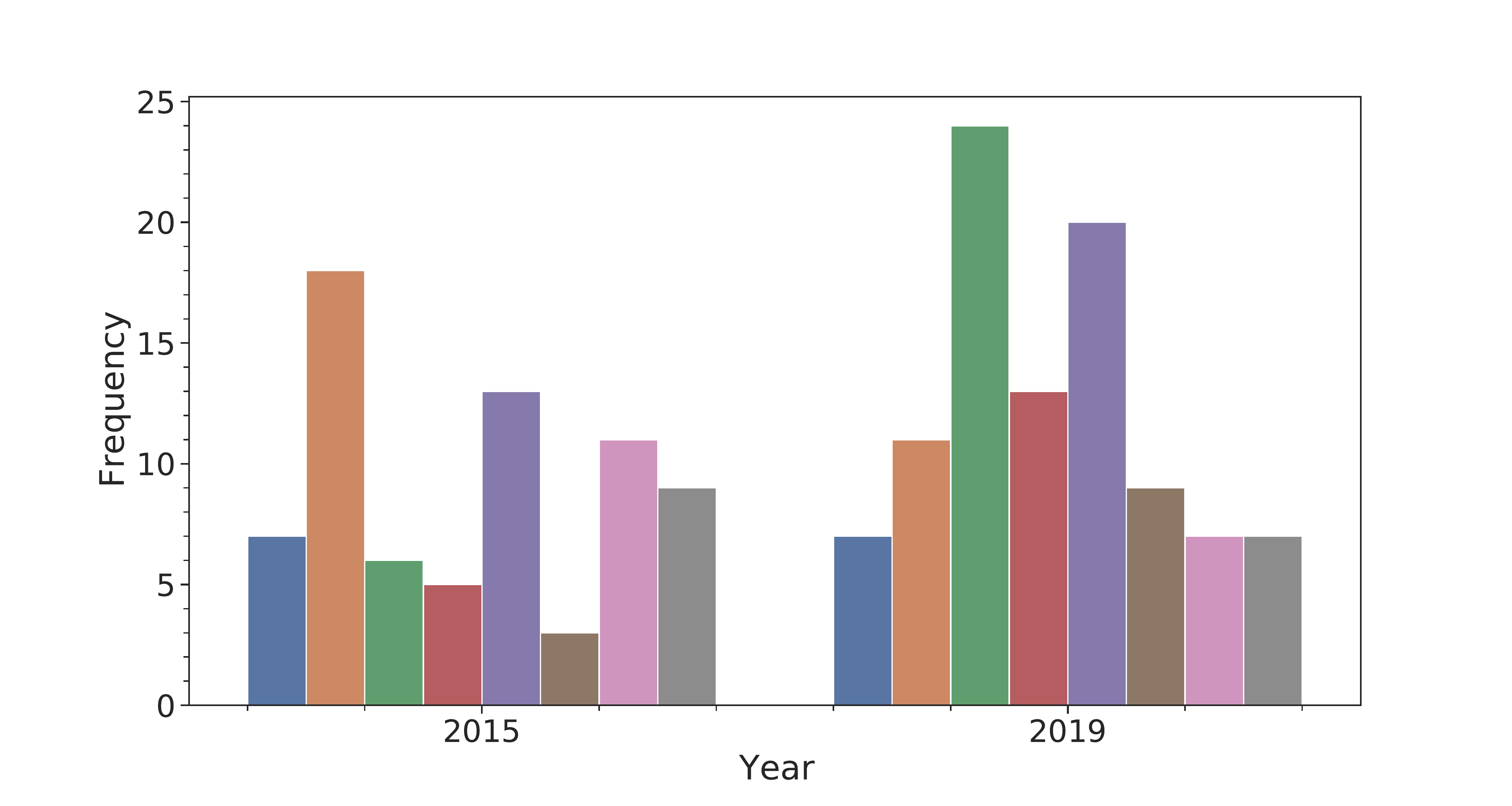}
    \subcaption{Reported materials in Korea}
    \label{fig:bar_Korea}
  \end{minipage}\\
  \begin{minipage}[b]{0.45\hsize}
    \centering
    \includegraphics[keepaspectratio, scale=0.15]{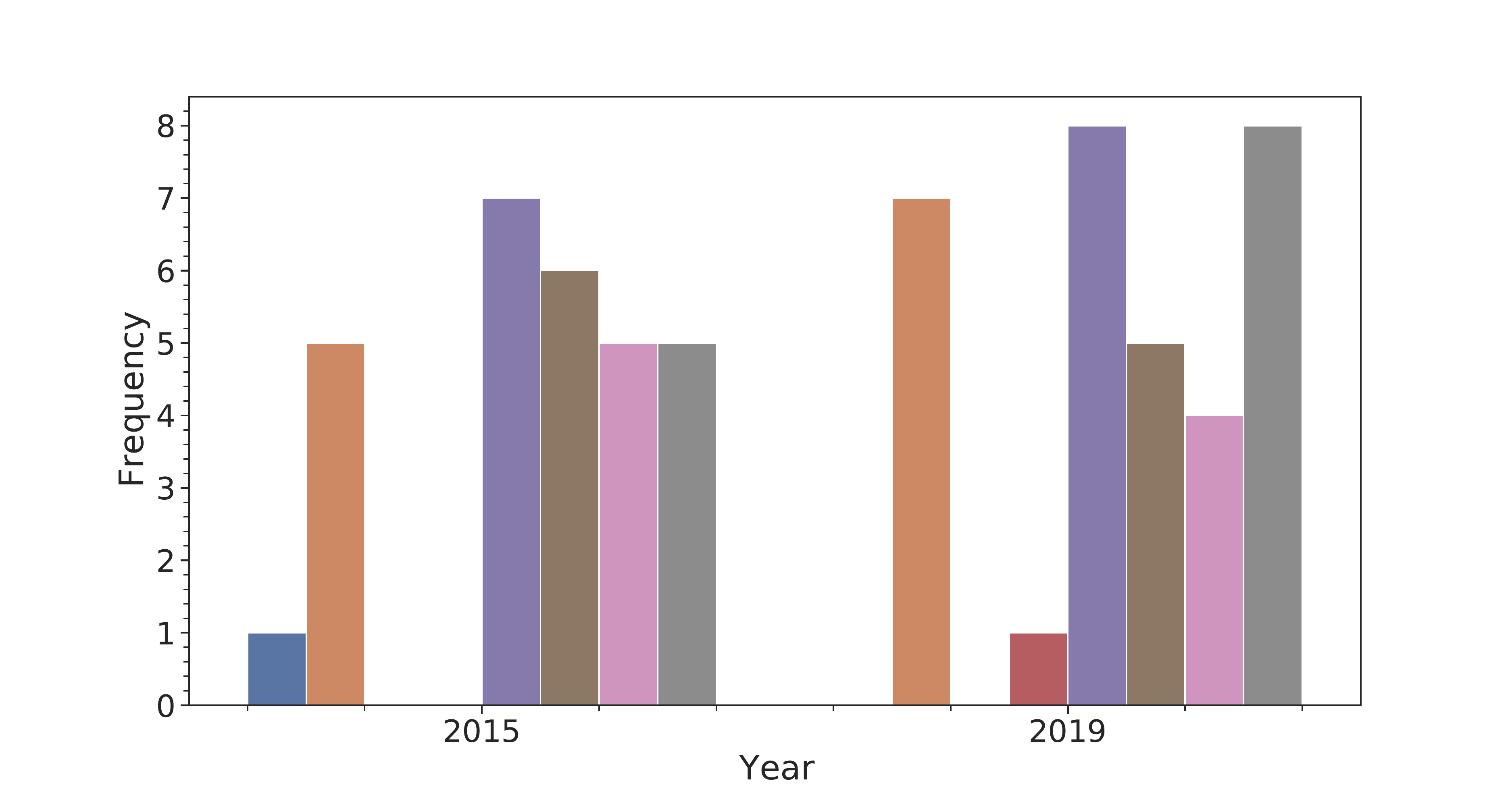}
    \subcaption{Reported materials in UK}
    \label{fig:bar_UK}
  \end{minipage}
  \begin{minipage}[b]{0.45\hsize}
    \centering
    \includegraphics[keepaspectratio, scale=0.15]{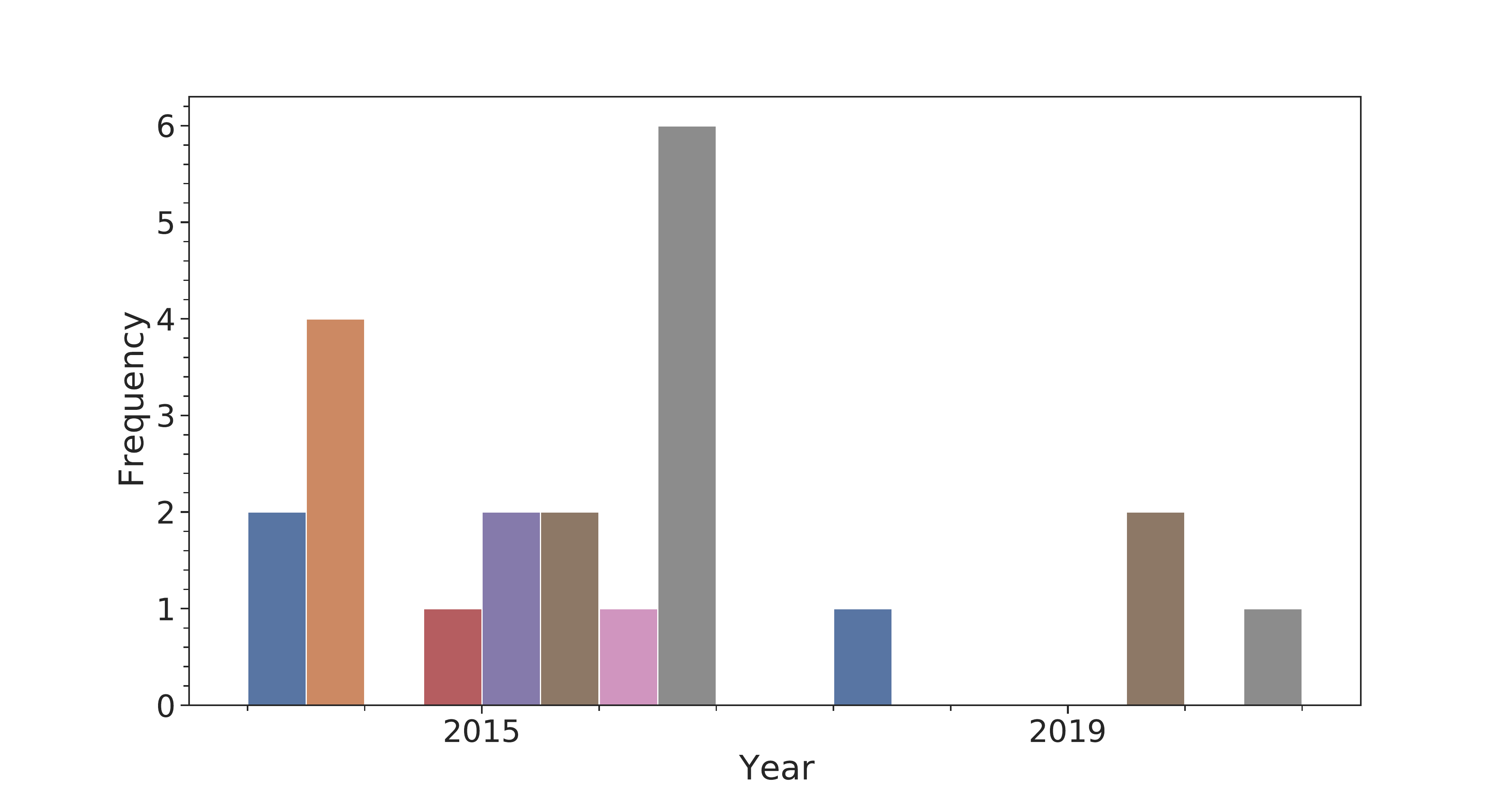}
    \subcaption{Reported materials in France}
    \label{fig:bar_France}
  \end{minipage}\\
  \begin{minipage}[b]{0.45\hsize}
    \centering
    \includegraphics[keepaspectratio, scale=0.15]{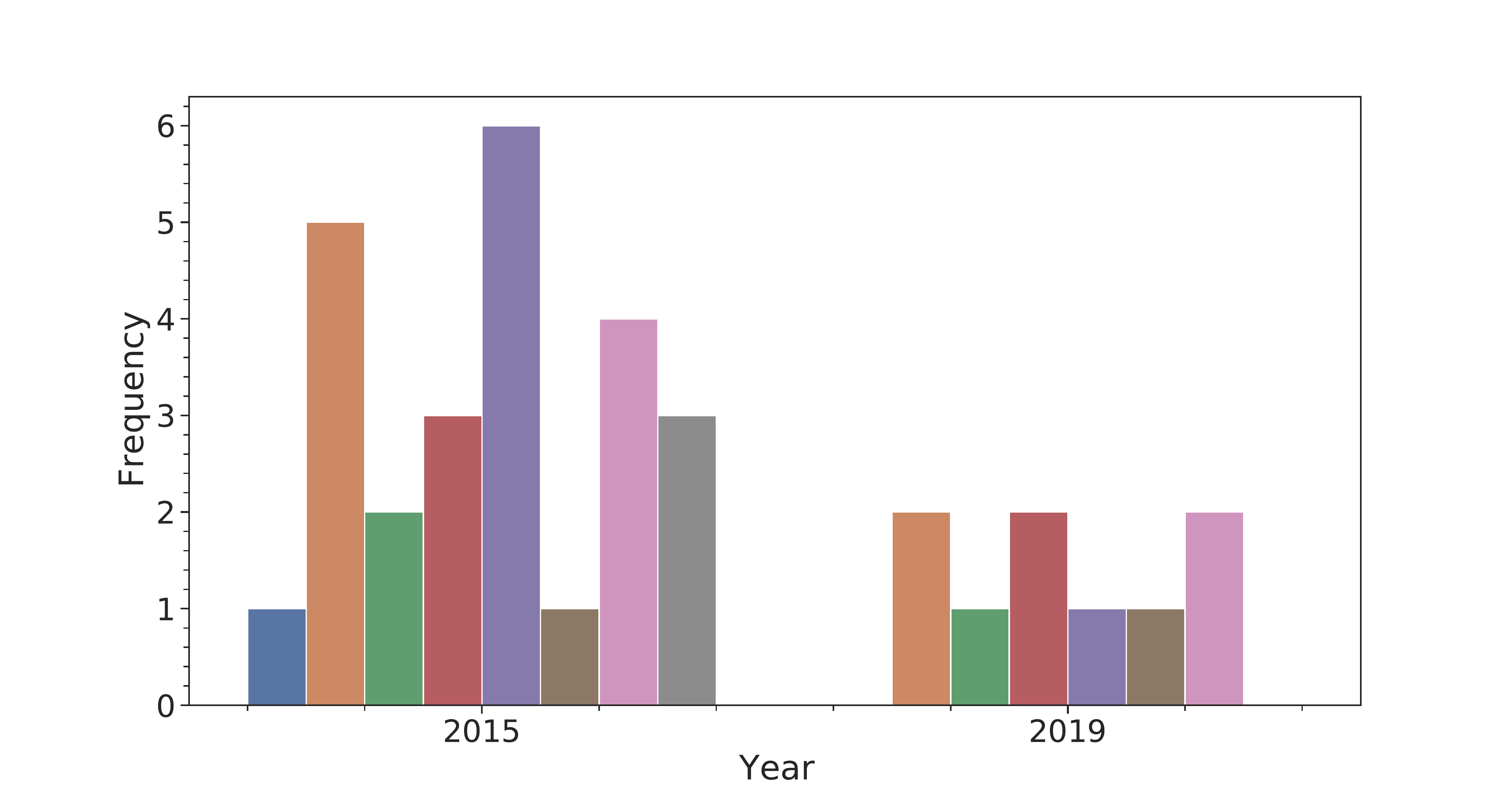}
    \subcaption{Reported materials in Taiwan}
    \label{fig:bar_Taiwan}
  \end{minipage}
  \begin{minipage}[b]{0.45\hsize}
    \centering
    \includegraphics[keepaspectratio, scale=0.15]{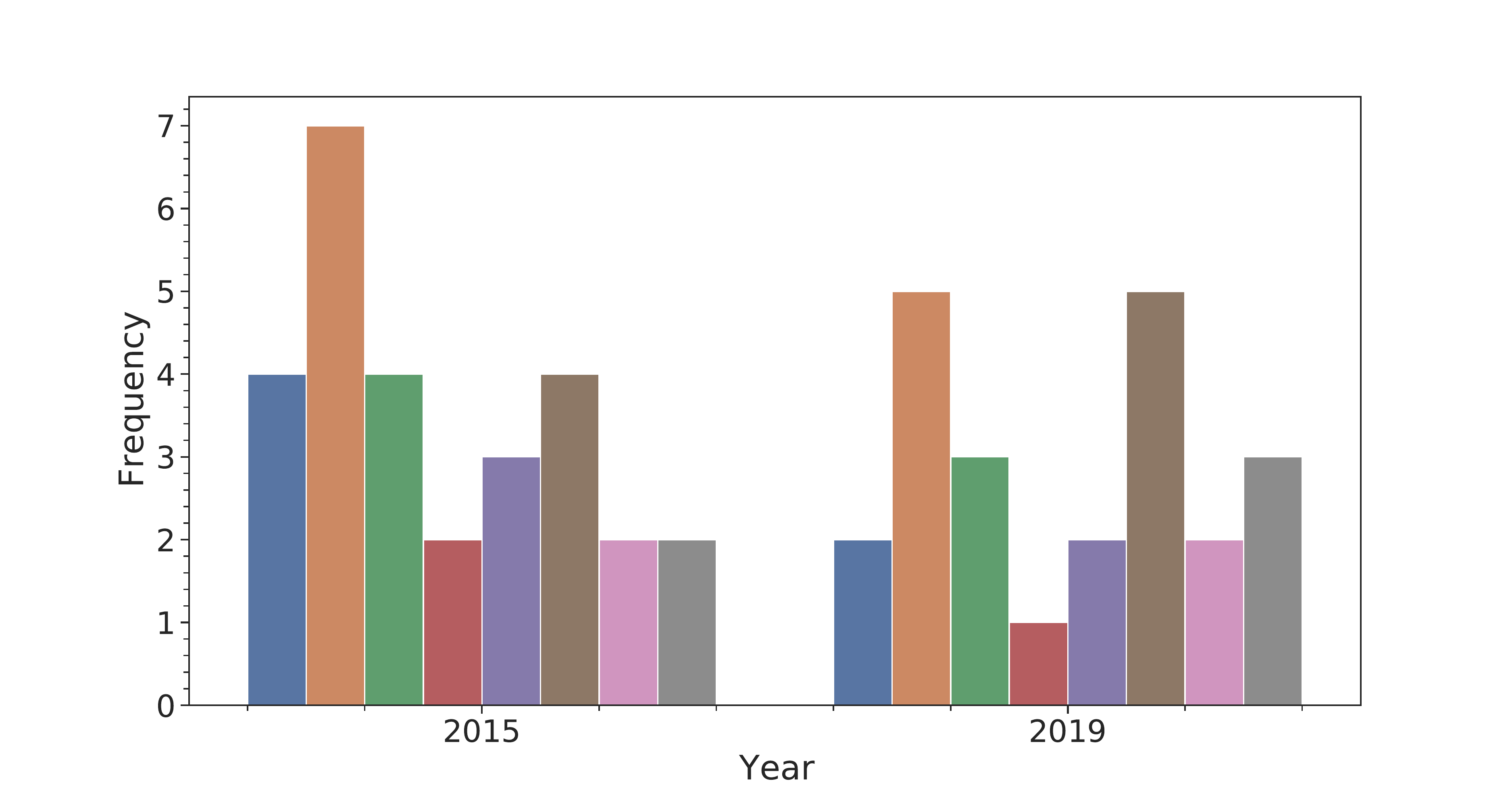}
    \subcaption{Reported materials in Germany}
    \label{fig:bar_Germany}
  \end{minipage}\\
  \caption{Year of transition of material by country.}
  \label{fig:bar_country}
\end{figure}

\textbf{Trend-by-country}: \Fref{fig:bar_country} is the year transition of material by six countries. It can be seen that the characteristics of the developed materials exist not only by year but also by country.

\begin{figure}[t]
  \begin{minipage}[b]{0.5\hsize}
    \centering
    \includegraphics[keepaspectratio, scale=0.18]{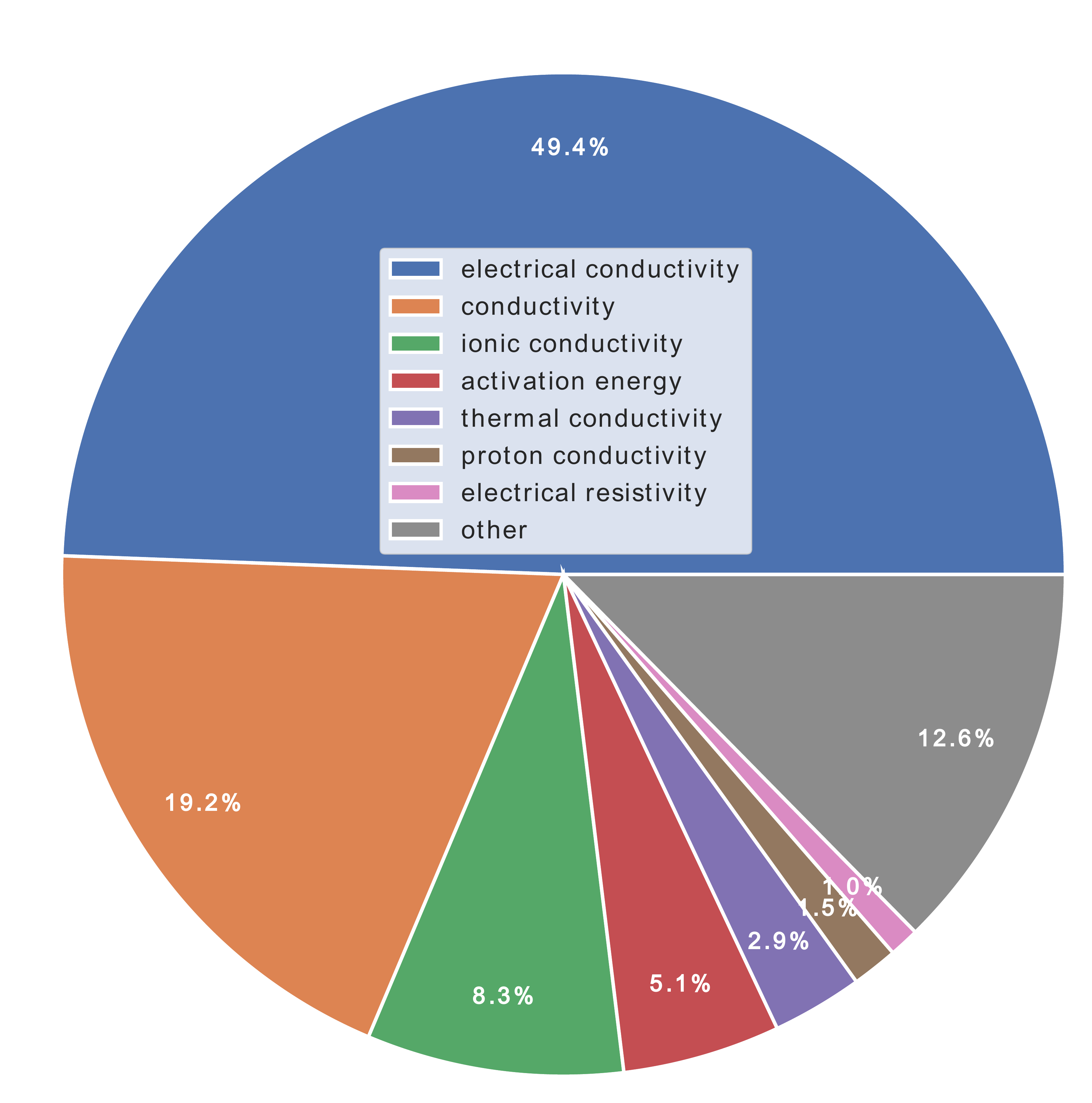}
    \subcaption{\characteristicname{} (19,644).}
    \label{fig:freq_characteristicname}
  \end{minipage}
  \begin{minipage}[b]{0.5\hsize}
    \centering
    \includegraphics[keepaspectratio, scale=0.18]{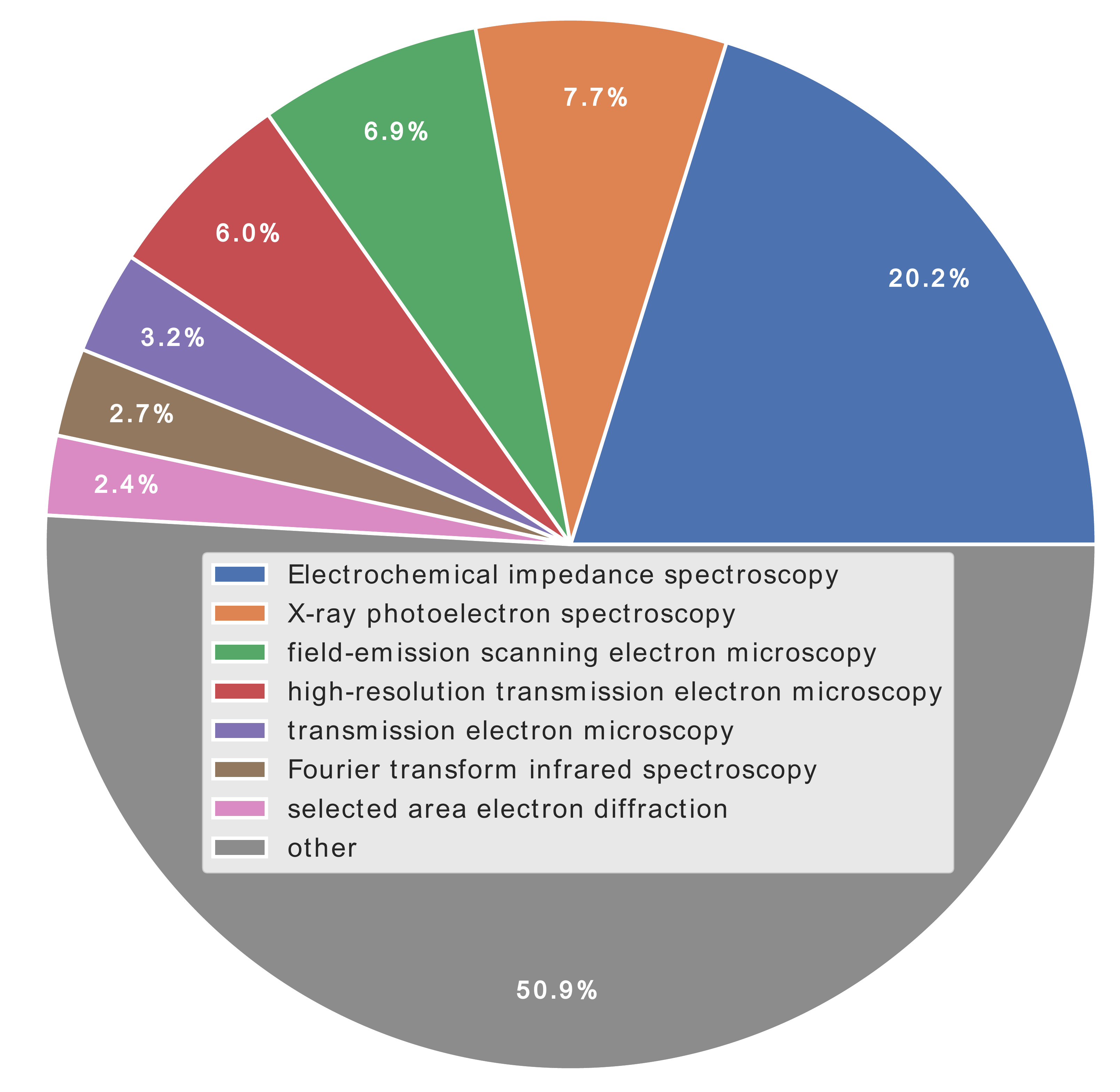}
    \subcaption{\propertymethod{} (41,150).}
    \label{fig:freq_method}
  \end{minipage}\\
  \begin{minipage}[b]{0.5\hsize}
    \centering
    \includegraphics[keepaspectratio, scale=0.18]{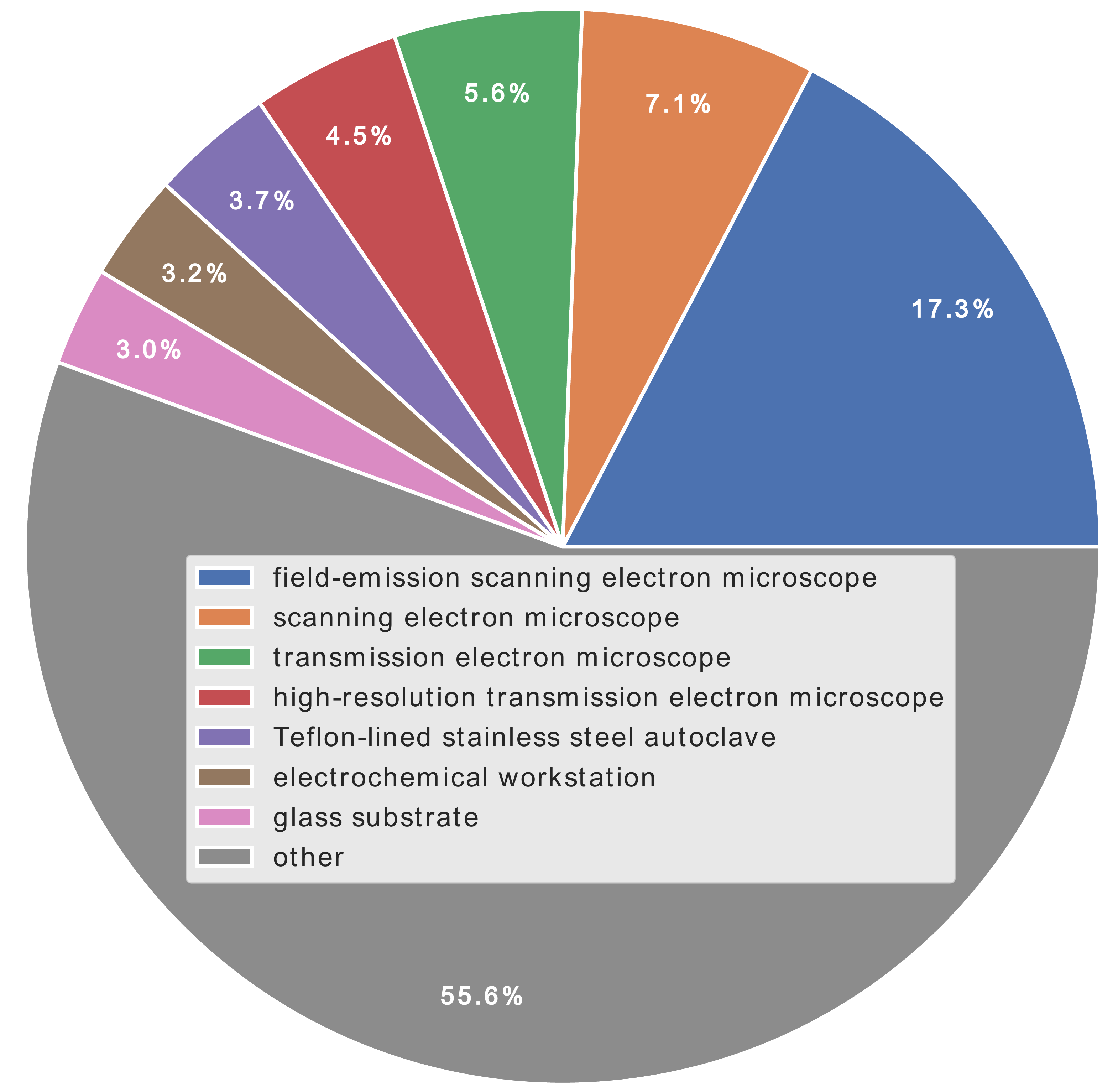}
    \subcaption{\propertyequipment{} (15,658).}
    \label{fig:freq_equipment}
  \end{minipage}
  \begin{minipage}[b]{0.5\hsize}
    \centering
    \includegraphics[keepaspectratio, scale=0.18]{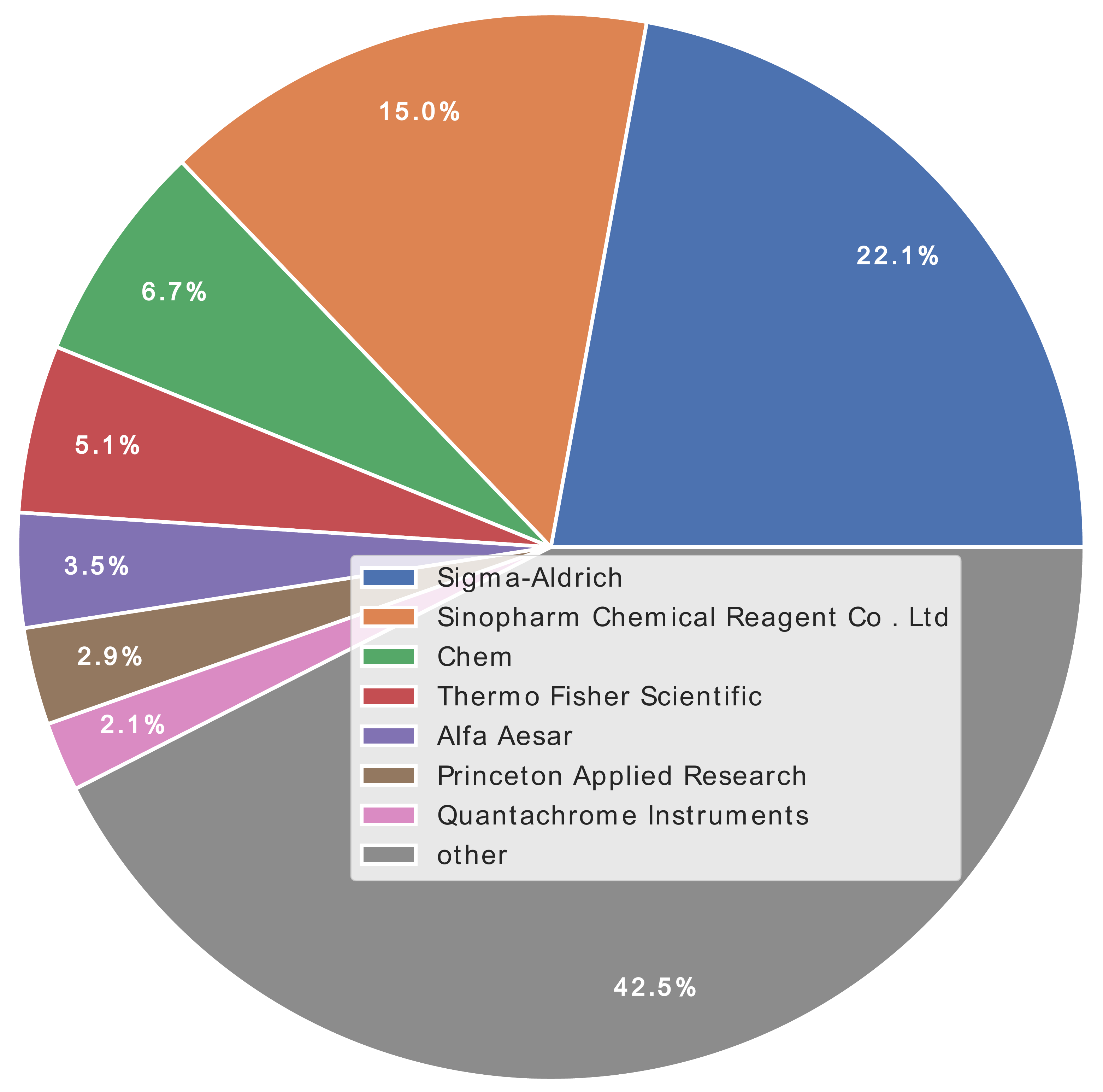}
    \subcaption{\propertymanufacturer{} (9,745).}
    \label{fig:freq_manufacturer}
  \end{minipage}
  \caption{Frequency of four properties in our automatically generated datasets: (a)\characteristicname{}, (b)\propertymethod{}, (c)\propertyequipment{}, (d)\propertymanufacturer{}. Number in brackets represents extracted phrases.}
  \label{fig:freq_properties}
\end{figure}

\textbf{Properties frequency}: \Fref{fig:freq_properties} shows frequency of four properties in our automatically generated datasets. There were many phrases related to ``conductivity'' extracted from \characteristicname{}, many phrases related to measurement extracted from \propertymethod{}, and many microscopes used for measurement extracted from \propertyequipment{}. In addition, \propertymanufacturer{} showed that among the material suppliers, ``Sigma-Aldrich'' and ``Sinopharm Chemical Reagent'' had a high share.


